\documentclass[11pt]{article}

\usepackage{acl}

\usepackage{times}
\usepackage{latexsym}
\usepackage{todonotes}

\usepackage[T1]{fontenc}
\usepackage[utf8]{inputenc}

\usepackage{microtype}
\usepackage{inconsolata}

\usepackage{graphicx}
\usepackage{booktabs}
\usepackage{multirow}
\usepackage{makecell}
\usepackage{colortbl}
\usepackage{subcaption}

\usepackage{amssymb}

\usepackage{enumitem}

\usepackage{mdframed}

\usepackage{amsmath,amsfonts,bm}

\def\eqref#1{equation~\ref{#1}}

\def\1{\bm{1}}

\DeclareMathAlphabet{\mathsfit}{\encodingdefault}{\sfdefault}{m}{sl}
\SetMathAlphabet{\mathsfit}{bold}{\encodingdefault}{\sfdefault}{bx}{n}

\definecolor{longctx}{RGB}{232,245,233}
\definecolor{simplerag}{RGB}{255,243,224}
\definecolor{embedrag}{RGB}{227,242,253}
\definecolor{structrag}{RGB}{243,229,245}
\definecolor{agentic}{RGB}{255,249,230}
\definecolor{promptbg}{gray}{0.95}

\title{Know It, Act on It: \\ Investigating Memory Utilization in LLM Personalization}

 \author{Zhaoxin Feng \and Jianfei Ma \and Emmanuele Chersoni \\
        The Hong Kong Polytechnic University\\
        \texttt{\{zhaoxinbetty.feng,jian-fei.ma\}@connect.polyu.hk,} \\ \texttt{emmanuele.chersoni@polyu.edu.hk}}
        
\begin{document}
\maketitle

\begin{abstract}
As large language model (LLM) agents evolve into personalized companions, memory has emerged as a core capability. However, LLMs face a knowledge utilization problem: they may fail to act on relevant user preferences even when they are fully present in context. When an agent fails to tailor its response in a context where previously shared user preferences should matter, it is unclear whether the model failed to remember that information or remembered it but failed to use it. To isolate this breakdown, we introduce a decoupled evaluation paradigm that administers paired \textit{Know} and \textit{Act} tests to the same user preference. We conduct large-scale experiments across 16 systems and five memory architectures, evaluating 1,000 preferences embedded at three levels of expression strength. Our results show a large gap between \textit{Know} and \textit{Act} outcomes: agents often pass the recall test for a user preference but fail to reflect that same preference in the paired behavioral scenario. While memory architectures reduce this gap, utilization remains especially weak for health and therapy-related preferences, where failures to act carry the greatest real-world stakes~\footnote{Code and data are available at \url{https://anonymous.4open.science/r/KnowAct_Benchmark-5BEB/}}.
\end{abstract}

\section{Introduction}
\label{sec:introduction}

\begin{figure}[t]
  \includegraphics[width=\columnwidth]{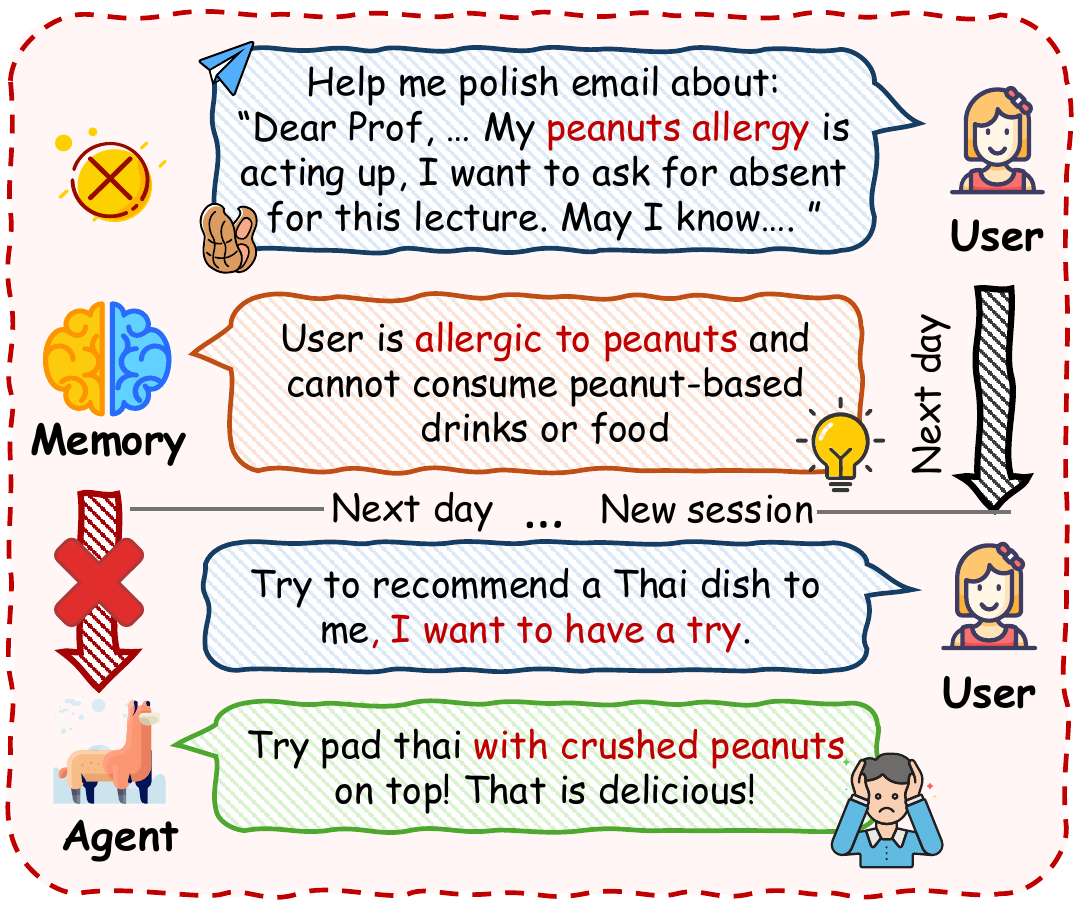}
  \caption{A memory-augmented LLM agent correctly stores a user's peanut allergy from an email-polishing request, yet fails to account for it when recommending a Thai dish in a subsequent session.}
  \label{fig:figure1}
\end{figure}

Large Language Model (LLM) agents are rapidly evolving from stateless tools into personalized long-term companions capable of sustained interaction with users over days, weeks, and months~\citep{openai2025personalized, meta2025superintelligence}. A core capability driving this transition is memory: the ability to store, organize, and retrieve information from past interactions~\citep{10.1145/3748302}. Both industry and academia are actively equipping agents with increasingly sophisticated memory architectures, from commercial systems such as ChatGPT's persistent memory~\citep{openai2024memory} and Claude's memory feature~\citep{anthropic2025memory} to open-source frameworks such as Mem0~\citep{chhikara2025mem0buildingproductionreadyai}, Letta/MemGPT~\citep{packer2024memgptllmsoperatingsystems}, and knowledge-graph-based systems like Zep~\citep{rasmussen2025zeptemporalknowledgegraph} and HippoRAG~\citep{gutierrez2025from}. These systems have demonstrated strong performance on benchmarks that test factual recall, multi-hop reasoning, and long-range understanding~\citep{wu2025longmemeval, hu2026evaluating}. But \emph{remembering} information is only the first step. LLMs face a well-documented knowledge utilization problem, often failing to act on relevant information even when it is fully present in context~\citep{shi2023large, liu-etal-2024-lost}. Memory modules could in principle alleviate this by distilling conversation histories into condensed, high-signal context, yet they also introduce new failure points such as semantic mismatch during retrieval. Whether memory-augmented LLM agents truly bridge the gap between remembering user preferences and acting on them remains an open question.

Most users do not explicitly state their preferences to chatbots. According to a recent large-scale analysis of ChatGPT usage~\citep{chatterji2025chatgpt}, the majority of users still treat LLMs as tools, asking them to refine emails, translate messages, or debug code, meaning their preferences are revealed only implicitly through everyday interactions~\citep{jiang2025personamemv2personalizedintelligencelearning}. As illustrated in Figure~\ref{fig:figure1}, a user asking a chatbot to polish an email might reveal a peanut allergy through the content. The memory module correctly extracts this preference. Yet in a later session, when the same user asks the agent to recommend a Thai dish, the agent enthusiastically suggests pad thai with crushed peanuts on top. Did the agent fail to \emph{remember} the allergy, or did it remember but fail to \emph{act} on it?

\begin{figure*}[t]
\centering
\includegraphics[width=1.0\linewidth]{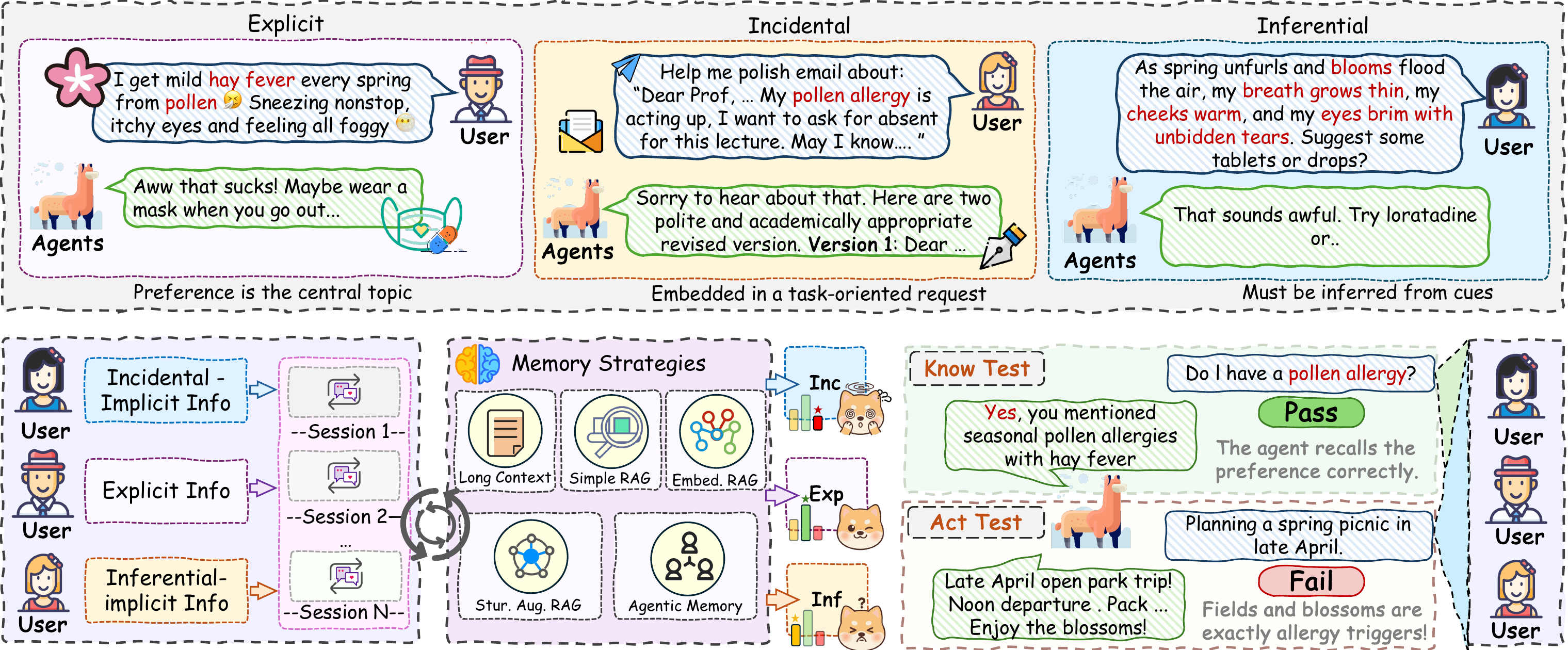}
\caption{\textbf{Top}: the same preference is expressed at three levels of strength during memory construction. \textbf{Bottom}: after incremental multi-session injection, a \textit{Know} test and an \textit{Act} test are administered independently on the same preference. Here the agent remembers the pollen allergy but fails to act on it when planning a spring outing.}
\label{fig:hero}
\end{figure*}

When such failures occur, determining their cause requires testing two capabilities \emph{separately on the same information}: whether the agent can recall it, and whether it can apply it in a relevant scenario. Testing only recall leaves open whether the information actually influences behavior; testing only behavior leaves open whether a good response reflects genuine memory or coincidence. Yet existing research, whether on LLM personalization or agentic memory, evaluates either recall or behavioral responses, but not both capabilities on the same stored information (\S\ref{sec:personalization}), and most studies operate in static long-context settings rather than exercising the incremental pipelines through which long-term agents' memory systems operate in practice. To bridge this gap, we design a paired experimental framework that applies a \textit{Know} test and an \textit{Act} test to \emph{the same} user preference. The \textit{Know} test directly asks whether the agent can recall a specific preference; the \textit{Act} test presents a natural scenario where that preference should influence behavior without mentioning it. We further introduce a three-level expression gradient (explicit, incidental, and inferential) extending the foundational explicit/implicit distinction~\citep{rich1979user} to investigate how the strength of preference expression affects downstream utilization. Following an incremental memory construction protocol~\citep{hu2026evaluating} that ensures each system processes information through its native pipeline, and building on PersonaMem-v2's persona and preference data~\citep{jiang2025personamemv2personalizedintelligencelearning}, we conduct large-scale experiments across 50 personas, 1,000 preferences, and three expression conditions, evaluating 16 memory systems spanning five architectural categories on 3,000 instances (see Figure~\ref{fig:hero}).

Our empirical investigation reveals a pervasive gap between \textit{Know} and \textit{Act} across 
modern LLM agents: regardless of whether a system relies on a pure 
long-context model or a dedicated memory architecture, the best 
performers convert no more than two-thirds of successfully remembered 
preferences into appropriate behavioral responses. Equipping LLM agents 
with memory systems substantially narrows this gap; Mem0~\citep{chhikara2025mem0buildingproductionreadyai}, for 
instance, raises GPT-4o-mini's utilization rate from 16.3\% to 
54.6\%, with gains disproportionately large on the behavioral side, 
suggesting that effective memory architectures contribute not merely 
by retrieving more, but by surfacing information in a form that the model 
can act upon. Across preference types, utilization diverges 
sharply even when recall is comparable: health and therapy related preferences consistently rank lowest, 
precisely the categories where failure to act carries the highest 
real-world stakes.

\section{Preliminaries}
\label{sec:preliminaries}

\subsection{LLM Personalization}
\label{sec:personalization}

Research on LLM personalization has developed along two largely disconnected tracks. One track deepens user understanding through person-level inference and multi-source reasoning~\citep{wu2026knowmebenchbenchmarkingpersonunderstanding, cheng2026lifebenchbenchmarklonghorizonmultisource}, yet evaluation remains question answering rather than behavioral. The other track measures behavior directly, testing whether agents generate tailored responses to explicit, implicit, or multi-session preferences~\citep{salemi-etal-2024-lamp, zhao2025do, li-etal-2025-toward, zollo2025personalllm, kumar2024longlampbenchmarkpersonalizedlongform}, maintain persona consistency~\citep{bosonai2024rpbench, wang-etal-2024-rolellm}, or apply preferences already provided in the prompt~\citep{yoon2026benchpresbenchmarkcontextawarepersonalized}. Each tests a different slice of the pipeline, yet none separately evaluates whether a preference is successfully remembered and whether it is also acted upon.
PersonaMem-v1~\citep{jiang2025personamem} is the only work that includes both knowledge and behavioral tests, but applies them to \emph{different} preferences, precluding paired diagnosis. PersonaMem-v2~\citep{jiang2025personamemv2personalizedintelligencelearning} adds implicit preferences and incremental construction but drops the knowledge test entirely.

\subsection{Memory Systems for LLM Agents}
\label{sec:memory}
As LLM-based agents transition from single-turn tools to persistent systems~\citep{xi2025rise, wang2024survey}, the ability to retain and utilize information across interactions has become a central design challenge~\citep{hu2026memoryageaiagents, huang2026rethinkingmemorymechanismsfoundation}. A growing body of research evaluates memory capabilities along increasingly sophisticated dimensions, from retrieval and comprehension in extended contexts~\citep{bai-etal-2024-longbench, zhang2024infinitebench, maharana-etal-2024-evaluating}, to multi-hop reasoning and temporal understanding~\citep{bai-etal-2025-longbench, wu2025longmemeval}, to memory reliability~\citep{chen2026halumemevaluatinghallucinationsmemory}, and recently to four integrated competencies evaluated through incremental input protocols~\citep{hu2026evaluating}. Despite this expansion, the entire line of work evaluates whether information can be \emph{recovered}, not whether it \emph{shapes downstream behavior}~\citep{hu2026memoryageaiagents, huang2026rethinkingmemorymechanismsfoundation}.

Existing memory architectures span five broad categories evaluated in our experiments. \textbf{Long-context baselines} such as GPT-4o-mini~\citep{openai2024gpt4omini}, GPT-4o~\citep{openai2024gpt4o}, Gemini 3.1 Flash~\citep{google2026gemini31pro}, and Claude 4.6 Sonnet~\citep{anthropic2026claude46} place the entire conversation history in the context window, eliminating retrieval errors but introducing attention-allocation bottlenecks~\citep{liu-etal-2024-lost}. \textbf{RAG-based methods} retrieve relevant passages at query time via sparse lexical matching (BM25~\citep{robertson1994some}) or dense vector retrieval (text-embedding-3-small/-large~\citep{openai2024textembedding}, Qwen3-Embedding-4B~\citep{qwen2025embedding}), but are limited to top-kk
k selection. \textbf{Structure-augmented RAG} organizes information into knowledge graphs or structured stores: Mem0~\citep{chhikara2025mem0buildingproductionreadyai} extracts atomic facts, Zep/Graphiti~\citep{rasmussen2025zeptemporalknowledgegraph} builds temporal knowledge graphs, Cognee~\citep{markovic2025optimizinginterfaceknowledgegraphs} refines its graph via feedback loops, and HippoRAG-v2~\citep{gutierrez2025from} uses hippocampus-inspired pattern completion. \textbf{Agentic memory} systems give the agent control over its own memory operations: Letta/MemGPT~\citep{packer2024memgptllmsoperatingsystems} manages hierarchical memory tiers, Self-RAG~\citep{asai2024selfrag} learns when to retrieve, MemoryOS~\citep{kang-etal-2025-memory} applies OS-inspired memory management, and A-MEM~\citep{xu2025amem} uses Zettelkasten-based note linking.

\section{Experimental Design}
\label{sec:knowact}

This section describes our experimental design, illustrated in Figure~\ref{fig:hero}: the data construction pipeline (\S\ref{sec:construction}) and the paired evaluation framework (\S\ref{sec:task}).

\subsection{Data Construction}
\label{sec:construction}


\textbf{Personas and Preferences.} From PersonaMem-v2's personas~\citep{jiang2025personamemv2personalizedintelligencelearning}, we select 50 via stratified sampling across 11 cultural regions, filtering for non-null nationality and sufficient preference count ($>$70). The full filtering pipeline is described in Appendix~\ref{app:persona-selection}. For each persona, we sample 20 preferences as \emph{target} preferences, which are the preferences that will be evaluated through paired \textit{Know} and \textit{Act} tests. The remaining preferences (approximately 50--70 per persona) serve as \emph{non-target} context: their conversation entries are injected alongside target preferences to provide realistic retrieval noise, reflecting the fact that real user--chatbot histories contain a mix of topics and not every conversation is relevant to a given query. Target preferences are sampled according to a fixed quota across five preference types (stereotypical, anti-stereotypical, neutral, health/medical, and therapy/emotional), with a fallback mechanism (Appendix~\ref{app:preference-sampling}).

\textbf{Expression Strength Gradient.} In everyday user--chatbot interactions, preferences are conveyed at varying levels of directness, from explicit statements to subtle behavioral cues~\citep{rich1979user, jiang2025personamemv2personalizedintelligencelearning}. We introduce a three-level gradient that captures this spectrum. For each preference, we construct three versions of the conversation snippet in which the preference is embedded, while keeping the \textit{Know} and \textit{Act} test questions identical across all three versions. GPT-5~\citep{openai2026gpt5} generates the explicit and inferential snippets (Appendix~\ref{app:snippet-prompts}); the incidental version uses PersonaMem-v2's original conversation entry directly, which was also generated by GPT-5~\citep{openai2026gpt5}. All three variants share identical non-target chunks, ensuring that performance differences reflect only expression strength.
\begin{itemize}[leftmargin=1.5em, itemsep=2pt]
    \item \textit{Explicit}: The user directly states the preference as the central topic of conversation (e.g., ``Just to let you know, every spring I get hay fever from pollen. It's mild, but I sneeze a lot and my eyes get itchy'').
    \item \textit{Incidental}: The preference appears as incidental information within a task-oriented conversation but is never directly stated; extracting it requires reading comprehension across the surrounding context (e.g., the user asks to polish an email that mentions sneezing often and meeting at a tea house instead of the park).
    \item \textit{Inferential}: The preference must be inferred from the user's behavioral patterns; no single utterance explicitly labels it. Although never directly stated, the conversation revolves around closely related topics, allowing LLM agents to deduce the preference from contextual cues (e.g., the user mentions visiting outdoor temples during a spring festival, describes sneezing and watery eyes on windy days with blossoms, and asks for tablets that will not make them sleepy, without ever mentioning `allergy').
\end{itemize}

Representative examples of all three expression levels and their corresponding \textit{Know}/\textit{Act} tests are provided in Appendix~\ref{app:examples}.

\textbf{Chunk Sequences and Test Pairs.} For each persona, we construct three parallel chunk sequences, one per expression level, resulting in 150 sequences (50 personas $\times$ 3 expression levels) comprising 11,955 total chunks and approximately 4.55 million tokens (target-to-non-target ratio $\approx$ 1:3). The three sequences for a given persona differ only in their 20 target preference snippets; all non-target chunks are shared. GPT-5~\citep{openai2026gpt5} generates a paired \textit{Know} test (question and ground truth answer) and \textit{Act} test (scenario) for each of the 1,000 preferences (Appendix~\ref{app:test-prompt}). The same test questions are used across all three expression levels, ensuring that performance differences reflect signal strength, not test difficulty. Quality validation details are provided in Appendix~\ref{app:quality}. Complete dataset statistics are in Appendix~\ref{app:data-details} (Table~\ref{tab:dataset-statistics-appendix}).

\subsection{Paired Know--Act Evaluation}
\label{sec:task}

The central design principle of our framework is that every user preference is evaluated through two complementary tests that target the same underlying information but probe different capabilities.

\textbf{Know Test.} The \textit{Know} test establishes whether the agent has successfully stored and can retrieve a specific user preference. Questions are phrased in first person (e.g., ``Do I have any seasonal allergies?'') to simulate a user asking about information shared in prior interactions. An LLM judge (GPT-5~\citep{openai2026gpt5}) compares the agent's response against a ground truth answer and assigns a binary Pass/Fail verdict (Appendix~\ref{app:know-judge}). Know questions are deliberately simple and direct: failure indicates unambiguous memory failure. To validate that these questions cannot be answered by prior knowledge or guessing alone, we evaluate all generation models used in our experiments on the full 1,000 Know questions without any conversation history. All four models score below 3\%, confirming that non-trivial \textit{Know} accuracy in memory-augmented systems genuinely reflects successful memorization (Appendix~\ref{app:zero-memory-know}).

\begin{table*}[htbp]
\centering
\caption{\textit{Know Accuracy}, \textit{Act Accuracy}, and \textit{Utilization Rate} across memory systems. Exp\,=\,Explicit, Inc\,=\,Incidental, Inf\,=\,Inferential. Avg\,=\,macro-average over the three conditions. Long-context baselines use their native models; all other systems use GPT-4o-mini as the generation backbone. \textbf{Bold}\,=\,best; \underline{underline}\,=\,second; \textit{italic}\,=\,third.}
\label{tab:main-results}
\fontsize{9}{11}\selectfont
\setlength{\tabcolsep}{3pt}
\begin{tabular}{ll rrrr rrrr rrrr}
\toprule
& & \multicolumn{4}{c}{\textit{Know Acc.}\ (\%)} & \multicolumn{4}{c}{\textit{Act Acc.}\ (\%)} & \multicolumn{4}{c}{\textit{Util.}\ \textit{Rate} (\%)} \\
\cmidrule(lr){3-6} \cmidrule(lr){7-10} \cmidrule(lr){11-14}
Type & System & Exp & Inc & Inf & Avg & Exp & Inc & Inf & Avg & Exp & Inc & Inf & Avg \\
\midrule
\rowcolor{longctx} & GPT-4o-mini & 83.7 & 9.1 & 11.3 & 34.7 & 19.8 & 14.4 & 17.6 & 17.3 & 20.0 & 12.1 & 16.8 & 16.3 \\
\rowcolor{longctx} & GPT-4o & \textit{99.0} & \textit{37.2} & \textit{60.7} & \textit{65.6} & 37.7 & 22.7 & 27.4 & 29.3 & 37.9 & 28.5 & 30.0 & 32.1 \\
\rowcolor{longctx} & Gemini 3.1 Flash & \textbf{99.5} & \textbf{56.1} & \textbf{80.9} & \textbf{78.8} & \underline{71.9} & \underline{37.6} & \underline{50.8} & \underline{53.4} & \textit{71.9} & \underline{47.1} & \underline{52.2} & \underline{57.0} \\
\rowcolor{longctx} \multirow{-4}{*}{Long-Context} & Claude 4.6 Sonnet & \underline{99.3} & \underline{49.0} & \underline{80.4} & \underline{76.2} & \textbf{82.2} & \textbf{42.7} & \textbf{59.5} & \textbf{61.5} & \textbf{82.2} & \textbf{50.8} & \textbf{62.3} & \textbf{65.1} \\
\midrule
\rowcolor{simplerag} Simple RAG & BM25 & 86.7 & 16.8 & 13.7 & 39.1 & 19.3 & 12.0 & 18.3 & 16.5 & 20.0 & 11.3 & 20.4 & 17.2 \\
\midrule
\rowcolor{embedrag} & RAG te3-small & 98.0 & 18.4 & 38.1 & 51.5 & 45.2 & 13.2 & 25.0 & 27.8 & 45.6 & 18.5 & 31.0 & 31.7 \\
\rowcolor{embedrag} & RAG te3-large & 97.6 & 20.0 & 39.7 & 52.4 & 44.5 & 13.2 & 25.3 & 27.7 & 45.0 & 19.0 & 31.5 & 31.8 \\
\rowcolor{embedrag} \multirow{-3}{*}{Embedding RAG} & RAG Qwen3-Emb-4B & 98.7 & 20.6 & 50.2 & 56.5 & 53.1 & 13.1 & 25.8 & 30.7 & 53.6 & 15.5 & 31.5 & 33.5 \\
\midrule
\rowcolor{structrag} & Mem0 & 97.7 & 30.9 & 52.3 & 60.3 & \textit{71.2} & \textit{27.0} & \textit{39.9} & \textit{46.0} & \underline{72.0} & \textit{45.6} & \textit{46.3} & \textit{54.6} \\
\rowcolor{structrag} & Zep/Graphiti & 52.9 & 2.3 & 2.5 & 19.2 & 11.0 & 10.8 & 11.7 & 11.2 & 12.3 & 21.7 & 16.0 & 16.7 \\
\rowcolor{structrag} & Cognee & 90.3 & 15.6 & 23.8 & 43.2 & 32.1 & 15.1 & 24.2 & 23.8 & 32.7 & 19.2 & 29.0 & 27.0 \\
\rowcolor{structrag} \multirow{-4}{*}{Structure-Aug.\ RAG} & HippoRAG-v2 & 98.0 & 16.2 & 26.6 & 46.9 & 35.1 & 15.0 & 24.0 & 24.7 & 35.3 & 21.0 & 27.8 & 28.0 \\
\midrule
\rowcolor{agentic} & Letta/MemGPT & 89.5 & 14.3 & 22.1 & 42.0 & 32.5 & 14.5 & 22.2 & 23.1 & 32.6 & 16.1 & 29.4 & 26.0 \\
\rowcolor{agentic} & Self-RAG & 94.2 & 25.0 & 53.8 & 57.7 & 61.4 & 9.2 & 27.2 & 32.6 & 62.1 & 10.0 & 33.8 & 35.3 \\
\rowcolor{agentic} & MemoryOS & 93.5 & 11.7 & 32.3 & 45.8 & 44.9 & 13.2 & 25.3 & 27.8 & 46.3 & 16.2 & 37.2 & 33.2 \\
\rowcolor{agentic} \multirow{-4}{*}{Agentic Memory} & A-MEM & 89.8 & 14.8 & 20.1 & 41.6 & 25.8 & 16.4 & 21.2 & 21.1 & 26.2 & 19.6 & 24.4 & 23.4 \\
\bottomrule
\end{tabular}
\end{table*}

\textbf{Act Test.} The \textit{Act} test evaluates whether stored knowledge \emph{influences behavior}. The agent is presented with a natural scenario where the preference should shape its response. The scenario itself does not reference the preference; the agent is expected to consider it based on what it has stored in memory. For instance, if the user has a pollen allergy, the Act test might ask the agent to ``suggest a location and packing list for a spring picnic,'' expecting the agent to consider allergen exposure in its recommendation. We adopt paired comparison scoring: the agent's response with memory is compared against a response from the same backbone model without any memory input. An LLM judge (GPT-5) determines whether the memory-augmented response better reflects consideration of the preference, producing a binary Pass/Fail outcome (Appendix~\ref{app:act-judge}).

\textbf{Evaluation Protocol.} Following the inject-then-query paradigm of \textit{MemoryAgentBench}~\citep{hu2026evaluating}, we feed conversation chunks to memory systems one at a time using a standardized memorization template (Appendix~\ref{app:injection-prompt}). This ensures that each system processes information through its native mechanisms: Mem0~\citep{chhikara2025mem0buildingproductionreadyai} performs fact extraction, Letta~\citep{packer2024memgptllmsoperatingsystems} manages hierarchical memory tiers, and knowledge-graph-based systems construct their graphs progressively. After all chunks are injected, the \textit{Know} and \textit{Act} tests are administered in separate, independent sessions to avoid priming effects. Long-context baselines receive all chunks concatenated into a single context window preceding the test query. All long-context baselines use temperature=0 for deterministic generation. All RAG and agentic memory systems use GPT-4o-mini as the shared generation backbone, ensuring that performance differences reflect memory architecture rather than generation capability. Systems requiring embedding-based retrieval use text-embedding-3-small~\citep{openai2024textembedding} as the default embedding model, with the exception of MemoryOS~\citep{kang-etal-2025-memory}, whose implementation hardcodes all-MiniLM-L6-v2~\citep{wang2020minilm}.

\section{Experiments}
\label{sec:experiments}

\subsection{Setup}
\label{sec:setup}

We evaluate 16 memory systems spanning all five categories of the taxonomy introduced in \S\ref{sec:memory}, covering long-context baselines, simple RAG, embedding-based RAG, structure-augmented RAG, and agentic memory. 

\begin{figure*}[hbtp]
\centering
\includegraphics[width=1.0\linewidth]{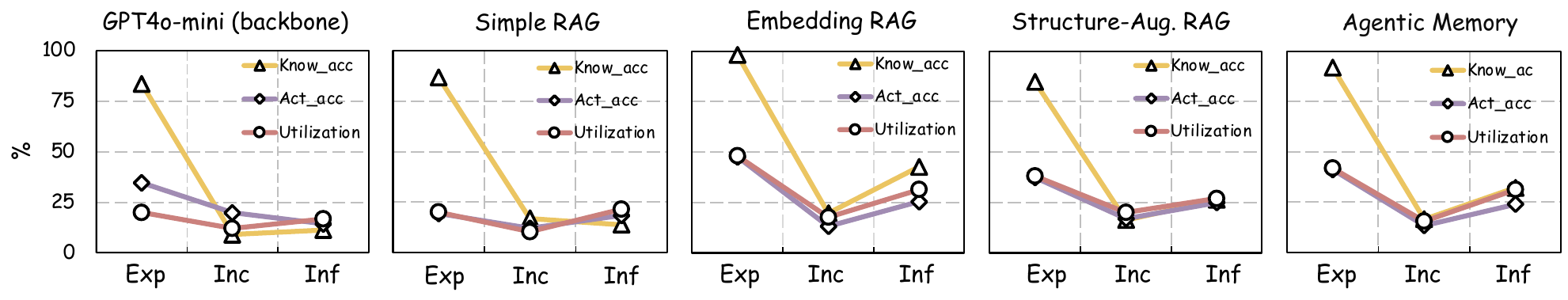}
\caption{\textit{Know Accuracy}, \textit{Act Accuracy}, and \textit{Utilization Rate} across expression levels for each category. Exp\,=\,Explicit, Inc\,=\,Incidental, Inf\,=\,Inferential. All architectures show a sharp \textit{Know} drop from Explicit to Incidental, with partial recovery under Inferential. Act and \textit{Utilization} remain comparatively flat.}
\label{fig:expression-level}
\end{figure*}

\textbf{Metrics.} For each preference, the \textit{Know} test and \textit{Act} test each produce a binary Pass/Fail outcome, yielding four possible states per preference: both pass (\textsc{kp\,ap}), \textit{Know} pass but \textit{Act} fail (\textsc{kp\,af}), \textit{Know} fail but \textit{Act} pass (\textsc{kf\,ap}), and both fail (\textsc{kf\,af}). We report \textit{Know Accuracy} and \textit{Act Accuracy} as the pass rates of each test, respectively. We additionally report the \textit{Utilization Rate}, defined as the fraction of correctly remembered preferences that the agent also acts on:
\[
\text{\textit{Utilization Rate}} = \frac{N_{\textsc{kp,ap}}}{N_{\textsc{kp,ap}} + N_{\textsc{kp,af}}}
\]
A system with high \textit{Know Accuracy} but low \textit{Utilization Rate} successfully retrieves user information yet fails to leverage it in downstream behavior, pointing to a utilization bottleneck rather than a memory failure.

\textbf{Judge Reliability.} Human evaluation of 100 \textit{Know} and 100 \textit{Act} judgments yields Fleiss' $\kappa$ of 0.89 and 0.82, respectively. Inter-judge agreement between GPT-5 and Claude Opus 4.6~\citep{anthropic2026claude46} on a separate 200-sample subset yields Cohen's $\kappa$ of 0.91 (\textit{Know}) and 0.88 (\textit{Act}), indicating strong agreement. Details are in Appendix~\ref{app:judge-reliability}.


\subsection{Overall Results}
\label{sec:results}
 
Table~\ref{tab:main-results} presents \textit{Know Accuracy}, \textit{Act Accuracy}, and \textit{Utilization Rate} for all 16 systems. We report the following findings.

Remembering does not imply acting: Know--Act dissociation is pervasive across all architectures. The central finding of KnowAct is that high recall does not guarantee behavioral utilization. Under the Explicit condition, most systems achieve high \textit{Know Accuracy}, yet \textit{Utilization Rates} vary widely across systems. Even the best-performing system, Claude 4.6 Sonnet, converts only about two-thirds of its remembered preferences into behavioral responses. The dissociation is most striking for GPT-4o-mini, which can articulate preferences when asked directly but rarely incorporates them into its responses. We note that KnowAct's retrieval setting is relatively favorable, as each preference corresponds to a single target passage with limited competing evidence; the Know--Act dissociation would likely be even more pronounced in realistic scenarios with greater retrieval noise.

Long-context models with 1M-token windows lead overall, but context window size is not the sole factor. Claude 4.6 Sonnet and Gemini 3.1 Flash (both 1M windows) substantially outperform GPT-4o and GPT-4o-mini (both 128K) across all metrics. However, model capability plays an equally important role: GPT-4o roughly doubles GPT-4o-mini's scores despite sharing the same 128K window, and among two 1M models, Gemini leads on \textit{Know} while Claude leads on \textit{Utilization}.
 
Memory architectures improve upon the GPT-4o-mini backbone, with Mem0 yielding the largest gain. Since all RAG and agentic memory systems share GPT-4o-mini as backbone, comparing them against GPT-4o-mini in the long-context setting isolates the benefit from memory architectures. Most systems improve substantially, such as Mem0~\citep{chhikara2025mem0buildingproductionreadyai} roughly triples the \textit{Utilization Rate}, achieving results comparable to Gemini 3.1 Flash.
 

\subsection{Analysis by Expression Level}

Figure~\ref{fig:expression-level} shows how \textit{Know Accuracy}, \textit{Act Accuracy}, and \textit{Utilization Rate} vary across the three expression levels for each architecture category.

The Incidental condition is harder to recall than Inferential across all categories. In every architecture category, \textit{Know Accuracy} reaches its lowest under the Incidental condition, not Inferential. This is counterintuitive: Inferential requires reasoning from behavioral cues, which should be more demanding. However, Incidental preferences are embedded as side details within task-oriented conversations and never become the focus, making them difficult for both memory systems and long-context models to flag as noteworthy. Under the Inferential condition, by contrast, conversations revolve around the relevant topic, providing richer semantic cues for retrieval.
 
\begin{figure*}[t]
\centering
\includegraphics[width=0.95\linewidth]{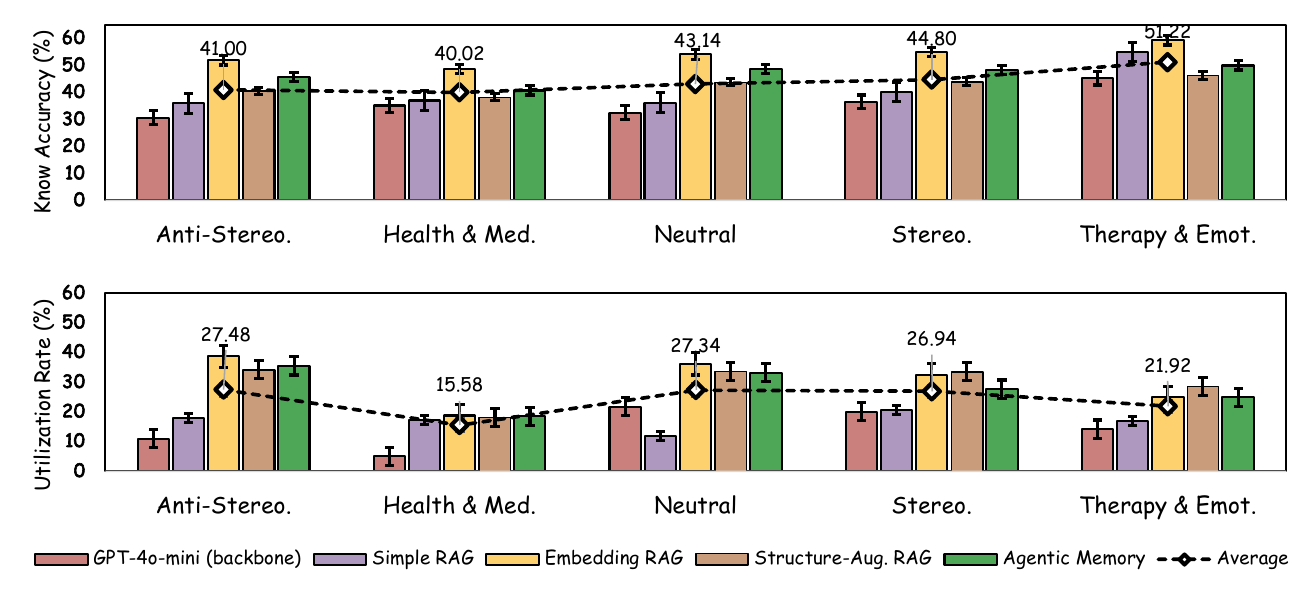}
\caption{\textit{Know Accuracy} (top) and \textit{Utilization Rate} (bottom), averaged by architecture category, split by preference type. Various memory systems backbone with GPT-4o-mini; long-context GPT-4o-mini results in Table~\ref{tab:main-results}. Therapy \& Emotional preferences are easy to recall but hard to utilize; Health \& Medical preferences score lowest on both metrics. Per-system details in Appendix~\ref{app:preference-details}.}
\label{fig:util-pref-type}
\vspace{-10pt}
\end{figure*}

Explicit expression exposes utilization failure; non-explicit expression exposes storage failure. Under the Explicit condition, \textit{Know Accuracy} is high across all architectures, yet \textit{Act Accuracy} and \textit{Utilization Rate} remain much lower, producing the largest Know-Act gap. This indicates that the primary bottleneck under Explicit expression is not memory but utilization: systems can recall the preference when asked directly but fail to incorporate it into behavioral responses.

\subsection{Analysis by Preference Type}
\label{sec:pref-analysis}

Figure~\ref{fig:util-pref-type} breaks down the\textit{ Utilization Rate} by preference type, with average values shown.

Health and therapy related preferences suffer the largest utilization gaps. \textit{Know Accuracy} is relatively uniform across preference types, indicating that storage difficulty alone does not explain performance differences. However, \textit{Utilization Rate} diverges sharply: Health \& Medical drops to the lowest average and Therapy \& Emotional is second-lowest, despite Therapy achieving the highest \textit{Know Accuracy}. This contrast indicates that the bottleneck lies in utilization rather than storage. These categories require more nuanced reasoning to translate into appropriate behavior: a pollen allergy should inform outdoor activity suggestions, and work-related stress should shape how the agent frames productivity advice, connections that are harder to make than simply recommending a cuisine matching a stated food preference. This has notable safety implications: unlike lifestyle preferences where commonsense reasoning often suffices, health and emotional preferences demand specialized domain knowledge to act upon correctly, and agents that retrieve yet fail to act on such information present a tangible risk in deployments where user wellbeing is at stake.
 
The GPT-4o-mini backbone relies more on population priors than memory-augmented systems. In the \textit{Utilization Rate} panel, GPT-4o-mini (backbone) performs better on Stereotypical than Anti-Stereotypical preferences, consistent with PersonaMem-v1~\citep{jiang2025personamem} and PersonaMem-v2~\citep{jiang2025personamemv2personalizedintelligencelearning} findings that models tend toward demographic expectations. This pattern reverses for memory-augmented systems, suggesting that retrieved memory overrides model priors, whereas the backbone falls back on prior-consistent behavior.

Memory architectures benefit utilization more than storage. Comparing memory-augmented systems against the GPT-4o-mini backbone, the relative gains in \textit{Utilization Rate} are consistently larger than those in \textit{Know Accuracy}. While \textit{Know Accuracy} improves modestly, \textit{Utilization Rate} roughly doubles across most preference types. This suggests that the primary value of memory architectures lies not only in improving what is retrieved, but in presenting information in a more actionable form that enables the backbone model to better incorporate preferences into its responses.

\subsection{Failure Attribution}
\label{sec:failure-attribution}

The analyses above reveal \textit{how much} Know--Act dissociation occurs but do not identify \textit{where} in the memory system pipeline the breakdown happens. An agent that knows a user's pollen allergy but recommends a spring picnic in an open field could be failing at any of several stages: the retrieval system may not have surfaced the allergy information for the activity recommendation query; the information may have been retrieved but the agent could not extract the relevant constraint from the surrounding conversational context; or the constraint may have been fully available but the agent still failed to incorporate it into its response. 

To localize failures, we conduct a controlled oracle study on Mem0~\citep{chhikara2025mem0buildingproductionreadyai}, the strongest non-long-context system, analyzing all 722 \textit{Know} Pass $\times$ \textit{Act} Fail cases across the three expression conditions.

\textbf{Method.} We design two oracle experiments that progressively bypass stages of the agentic memory pipeline. Both conditions use the same backbone model (GPT-4o-mini) and \textit{Act} judge as evaluation:
\begin{itemize}[leftmargin=*, topsep=2pt, itemsep=2pt, parsep=0pt]
    \item \textbf{Oracle chunk}: We bypass retrieval by injecting the ground-truth conversation chunk (the chunk containing the target preference) directly into the Act test context. The agent must still identify the relevant preference within it and apply it to the current task.
    \item \textbf{Oracle preference}: We bypass both retrieval and comprehension by injecting the preference text itself (e.g., ``This user has a pollen allergy'') directly into the Act test context. LLM agent's only remaining task is to apply the stated preference to the scenario.
\end{itemize}

Comparing outcomes across the three conditions yields a three-layer failure decomposition. For each case, exactly one attribution applies:
\begin{itemize}[leftmargin=*, topsep=2pt, itemsep=2pt, parsep=0pt]
    \item \textbf{Retrieval failure}: Oracle chunk passes, indicating that the original retrieval did not surface the relevant dialogue related to the target preference.
    \item \textbf{Comprehension failure}: Oracle chunk fails, but oracle preference passes, indicating that the relevant chunk was present but the agent could not extract the preference from it.
    \item \textbf{Application failure}: Oracle preference also fails, indicating that the preference was explicitly stated but the agent still did not act on it.
\end{itemize}

Across all conditions (see Table~\ref{tab:failure-attribution}), comprehension deficiency is the predominant cause of overall failures. More than half of the total errors occur because the model cannot accurately distill target features, even when the dialogue containing the target preference is already provided in the context. Furthermore, failures in the final application stage expose an intrinsic ceiling in translating known information into action. Regardless of how a preference is initially expressed, the proportion of application failures remains remarkably stable. In these cases, the model fails to follow the constraints despite being given explicit preference statements. This persistent inability to utilize provided knowledge reflects inherent limitations in the underlying language model regarding knowledge utilization.

\begin{table}[t]
\centering
\caption{Three-layer failure attribution for Mem0's Know Pass $\times$ Act Fail cases ($n{=}722$). Each case is classified into Retrieval~(R), Comprehension~(C), or Application~(A) failure. \textbf{Bold} marks the dominant failure type per condition.}
\label{tab:failure-attribution}
\begin{tabular}{lcrrr}
\toprule
Expression & $n$ & R (\%) & C (\%) & A (\%) \\
\midrule
Explicit     & 273 & \textbf{48.7} & 34.8 & 16.5 \\
Incidental   & 168 & 11.3 & \textbf{69.6} & 19.0 \\
Inferential  & 281 & 24.9 & \textbf{60.9} & 14.2 \\
\midrule
All          & 722 & 30.7 & \textbf{53.0} & 16.2 \\
\bottomrule
\end{tabular}
\end{table}

Varying expression strengths reveal distinct failure patterns. When preferences are explicitly stated, a counterintuitive retrieval bottleneck occurs, causing nearly half of the errors in this condition. Specifically, for an incoming query, the system fails to retrieve the exact conversation chunk containing the target preference that it is supposed to find. Conversely, for incidental or inferential preferences, the bottleneck shifts heavily to comprehension. In these non explicit scenarios, comprehension breakdowns dominate. The system struggles to extract hidden preferences from peripheral details or behavioral cues, even when directly provided with the relevant dialogue snippet.

\section{Conclusion}
\label{sec:conclusion}

This paper examines whether memory-augmented LLMs and long-context baselines can translate recalled preferences into appropriate behavior. Evaluating 16 systems across 1,000 preferences and three expression levels, we find that agents often achieve high accuracy on the \textit{Know} test but perform poorly on the \textit{Act} test. This gap varies by expression strength and preference type, with health and emotional domains proving especially challenging. Failure attribution further indicates that errors arise from retrieval, comprehension, and application, with comprehension failures dominating overall. These findings suggest that future personalization systems should move beyond simply storing and recalling user information, and instead focus on ensuring that personal memories are selectively retrieved, correctly interpreted, and reliably translated into context-appropriate behavior.

\section*{Limitations}

The evaluation relies on LLM judges, which may introduce systemic evaluation biases such as verbosity preference or insensitivity to subtle behavioral cues. Furthermore, our benchmark utilizes synthetic personas from PersonaMem-v2~\citep{jiang2025personamemv2personalizedintelligencelearning}; evaluating these systems on real-world, longitudinal user histories that feature natural topic drift and contradictions would strengthen ecological validity. Crucially, our design isolates the utilization bottleneck by linking each preference to a single target passage with limited competing distractors, inherently simplifying the retrieval task. Because real-world user interactions are vastly noisier and more fragmented, the pervasive Know-Act gap we observe is likely a conservative lower bound, one that would widen significantly in realistic deployment settings.


\section*{Ethics Statement}
We uses synthetic personas and preferences derived from PersonaMem-v2. All memory systems are evaluated in controlled settings without real user data. And we highlights critical safety concerns, particularly the utilization gap for health and medical information, which should inform the development of more reliable memory-enabled agents.


\bibliography{custom}

\appendix
\section{Memory System Implementation Details}
\label{app:system-details}

This section provides detailed technical descriptions and hyperparameter settings for each evaluated memory system. Table~\ref{tab:systems-appendix} provides an overview of all systems organized by the five-category taxonomy. All RAG and agentic memory systems use GPT-4o-mini (temperature=0) as the generation backbone.

\begin{table*}[htbp]
\centering
\caption{Memory systems evaluated in KnowAct, organized by the five-category taxonomy.}
\label{tab:systems-appendix}
\small
\setlength{\tabcolsep}{3pt}
\begin{tabular}{@{}p{2.5cm}lp{4.2cm}@{}}
\toprule
\textbf{Category} & \textbf{System} & \textbf{Key Mechanism} \\
\midrule
\multirow{4}{*}{Long-Context} & GPT-4o-mini~\citep{openai2024gpt4omini} & Full concatenation (128K) \\
 & GPT-4o~\citep{openai2024gpt4o} & Full concatenation (128K) \\
 & Gemini 3.1 Flash~\citep{google2026gemini31pro} & Full concatenation (1M) \\
 & Claude 4.6 Sonnet~\citep{anthropic2026claude46} & Full concatenation (1M) \\
\midrule
Simple RAG & BM25~\citep{robertson1994some} & Sparse lexical retrieval \\
\midrule
\multirow{3}{*}{\makecell[l]{Embedding\\RAG}} & RAG te3-small~\citep{openai2024textembedding} & Dense vector retrieval (1536d) \\
 & RAG te3-large~\citep{openai2024textembedding} & Dense vector retrieval (3072d) \\
 & RAG Qwen3-Emb-4B~\citep{qwen2025embedding} & Dense vector retrieval (1024d) \\
\midrule
\multirow{4}{*}{\makecell[l]{Structure-Aug.\\RAG}} & Mem0~\citep{chhikara2025mem0buildingproductionreadyai} & Fact extraction + vector/graph store \\
 & Zep/Graphiti~\citep{rasmussen2025zeptemporalknowledgegraph} & Temporal knowledge graph \\
 & Cognee~\citep{markovic2025optimizinginterfaceknowledgegraphs} & Feedback-refined KG \\
 & HippoRAG-v2~\citep{gutierrez2025from} & CLS-inspired pattern completion \\
\midrule
\multirow{4}{*}{\makecell[l]{Agentic\\Memory}} & Letta/MemGPT~\citep{packer2024memgptllmsoperatingsystems} & Hierarchical memory mgmt. \\
 & Self-RAG~\citep{asai2024selfrag} & Learned retrieval decisions \\
 & MemoryOS~\citep{kang-etal-2025-memory} & OS-inspired memory mgmt. \\
 & A-MEM~\citep{xu2025amem} & Zettelkasten-based memory \\
\bottomrule
\end{tabular}
\end{table*}

\subsection{Long-Context Baselines}

Long-context baselines receive all conversation chunks concatenated into a single context window preceding the test query. No retrieval or memory management is performed; the model attends over the full history directly.

\textbf{GPT-4o-mini}~\citep{openai2024gpt4omini} is used with a 128K context window. It serves as the shared generation backbone for all non-long-context systems, ensuring that performance differences across RAG and agentic systems reflect memory architecture rather than generation capability.

\textbf{GPT-4o}~\citep{openai2024gpt4o} is used with a 128K context window. Compared to GPT-4o-mini, it offers stronger reasoning capabilities at higher computational cost, allowing us to assess whether generation quality affects the Know--Act gap.

\textbf{Gemini 3.1 Flash}~\citep{google2026gemini31pro} supports a context window exceeding 1 million tokens, meaning the entire conversation history fits within the window for all personas in our benchmark. 

\textbf{Claude 4.6 Sonnet}~\citep{anthropic2026claude46} is evaluated with a 1M context window. As one of the strongest available models, it provides an upper-bound reference for what long-context processing can achieve on the KnowAct tasks.

\subsection{Simple RAG}

\textbf{BM25}~\citep{robertson1994some} uses the BM25 algorithm with default parameters ($k_1$=1.5, $b$=0.75) to rank chunks based on sparse lexical overlap. At query time, the top-10 chunks ranked by BM25 score are retrieved and concatenated as context for GPT-4o-mini.

\subsection{Embedding-based RAG}

We evaluate three embedding models of varying capacity, all using FAISS~\citep{johnson2019billion} for vector indexing and cosine similarity for retrieval. The top-10 most similar chunks are retrieved as context for GPT-4o-mini.

\textbf{RAG te3-small} uses OpenAI's text-embedding-3-small~\citep{openai2024textembedding} (1,536 dimensions), a lightweight embedding model suitable for high-throughput applications.

\textbf{RAG te3-large} uses OpenAI's text-embedding-3-large~\citep{openai2024textembedding} (3,072 dimensions), offering higher retrieval precision at increased computational cost.

\textbf{RAG Qwen3-Emb-4B} uses Qwen3-Embedding-4B~\citep{qwen2025embedding} (1,024 dimensions), a recent open-source embedding model that achieves competitive retrieval performance with a smaller dimensionality.

\subsection{Structure-Augmented RAG}

These systems augment retrieval with structured representations such as fact triples, knowledge graphs, or temporal indices. Each processes conversation chunks through its native ingestion pipeline during the injection phase.

\textbf{Mem0}~\citep{chhikara2025mem0buildingproductionreadyai} (v1.0.7) uses GPT-4o-mini to extract atomic facts from each conversation chunk during injection. Extracted facts are stored in both a Qdrant vector database and a knowledge graph. At query time, Mem0 retrieves relevant facts by combining vector similarity search with graph traversal.

\textbf{Zep/Graphiti}~\citep{rasmussen2025zeptemporalknowledgegraph} constructs a bi-temporal knowledge graph during chunk injection, tracking both when events occurred and when they were ingested. Entity and relationship extraction builds the graph incrementally. Retrieval combines graph traversal with semantic search over temporal subgraphs.

\textbf{Cognee}~\citep{markovic2025optimizinginterfaceknowledgegraphs} builds a knowledge graph using its Extract-Cognify-Load (ECL) pipeline. Each injected chunk updates the graph structure, and the system iteratively refines edge weights through feedback loops. Retrieval uses graph-augmented semantic search.

\textbf{HippoRAG-v2}~\citep{gutierrez2025from} implements a dual-system architecture inspired by complementary learning systems theory. An LLM serves as the neocortex, a knowledge graph as the hippocampal index, and personalized PageRank enables schema-based pattern completion for associative retrieval beyond direct keyword or embedding matching.

\subsection{Agentic Memory}

Agentic memory systems grant the agent control over its own memory operations, including what to store, when to retrieve, and how to organize accumulated knowledge.

\textbf{Letta/MemGPT}~\citep{packer2024memgptllmsoperatingsystems} manages a hierarchical memory system with three tiers: core memory (always present in context), recall memory (conversation history retrieved on demand), and archival memory (long-term storage for persistent facts). The agent autonomously issues function calls to move information between tiers, deciding what to retain in limited core memory and what to archive.

\textbf{Self-RAG}~\citep{asai2024selfrag} implements a multi-step self-reflective retrieval pipeline. The system first decides whether retrieval is needed for a given query, then retrieves relevant chunks, assesses their relevance, generates a candidate response, verifies whether the response is grounded in retrieved content, evaluates utility, and selects the best response. This process gives the agent autonomous control over when and what to retrieve.

\textbf{MemoryOS}~\citep{kang-etal-2025-memory} draws an analogy between memory management and operating system design, implementing memory allocation, paging, and garbage collection mechanisms. It uses all-MiniLM-L6-v2~\citep{wang2020minilm} as its embedding model, as specified in the original implementation, rather than the shared text-embedding-3-small used by other systems.

\textbf{A-MEM}~\citep{xu2025amem} organizes memory around the Zettelkasten (slip-box) method. Each new piece of information triggers the creation of a structured note with contextual descriptions, keywords, and tags. Critically, the system updates older notes in light of new information, enabling memory evolution rather than append-only accumulation.

\section{Prompt Templates and Judge Design}
\label{app:prompts}

\subsection{Memory Injection Prompt}
\label{app:injection-prompt}

For long-context agents, conversation chunks are concatenated using the MemoryAgentBench protocol~\citep{hu2026evaluating}. Each chunk is wrapped in a memorization template:

\begin{mdframed}[backgroundcolor=promptbg, linecolor=promptbg, skipabove=5pt, skipbelow=5pt, innerleftmargin=8pt, innerrightmargin=8pt, innertopmargin=8pt, innerbottommargin=8pt]
\small\ttfamily
<User> The following context is the conversation between the user and the assistant:\\
User: \{message\_1\}\\
Assistant: \{message\_2\}\\
...\\
<Assistant> I have memorized the conversation and I will answer the question you ask.
\end{mdframed}

For RAG-based systems, retrieved chunks are formatted as numbered memory entries:

\begin{mdframed}[backgroundcolor=promptbg, linecolor=promptbg, skipabove=5pt, skipbelow=5pt, innerleftmargin=8pt, innerrightmargin=8pt, innertopmargin=8pt, innerbottommargin=8pt]
\small\ttfamily
Memory 1:\\
\{chunk\_text\_1\}

Memory 2:\\
\{chunk\_text\_2\}

...

\{question\}
\end{mdframed}

\noindent where \texttt{\{message\_1\}}, \texttt{\{message\_2\}}, etc., are individual conversation turns. For RAG systems, \texttt{\{chunk\_text\_i\}} represents the retrieved chunk text, and \texttt{\{question\}} is the user's query.

\subsection{Know and Act Test Prompts}

Both Know and Act tests use a minimal system prompt to isolate the memory system's contribution:

\begin{mdframed}[backgroundcolor=promptbg, linecolor=promptbg, skipabove=5pt, skipbelow=5pt, innerleftmargin=8pt, innerrightmargin=8pt, innertopmargin=8pt, innerbottommargin=8pt]
\small\ttfamily
System: You are a helpful personal assistant.

User: \{know\_question or act\_scenario\}
\end{mdframed}

\noindent where \texttt{\{know\_question\}} is phrased in first person (e.g., ``Do I have any seasonal allergies?'') for Know tests, and \texttt{\{act\_scenario\}} presents a natural request without referencing the target preference (e.g., ``Can you recommend some fun weekend activities for this spring?'') for Act tests.

\subsection{Know Judge Prompt}
\label{app:know-judge}

The Know judge (GPT-5, reasoning effort=low) evaluates whether the agent correctly recalled a specific user preference:

\begin{mdframed}[backgroundcolor=promptbg, linecolor=promptbg, skipabove=5pt, skipbelow=5pt, innerleftmargin=8pt, innerrightmargin=8pt, innertopmargin=8pt, innerbottommargin=8pt]
\small\ttfamily
You are an impartial judge evaluating whether an AI assistant correctly recalled a specific user preference.

\textbf{Input}\\
- Question: The question asked to the assistant about the user's preference.\\
- Ground Truth: The correct answer about the user's preference.\\
- Assistant Response: The assistant's actual response.

\textbf{Evaluation Criteria}\\
Rate the response as Pass or Fail:

- Pass: The response conveys the same core information as the ground truth. Exact wording need not match, paraphrasing, reordering, or adding minor context is fine, as long as the essential preference is correctly identified.

- Fail: The response does not demonstrate knowledge of the specific preference.

\textbf{Output Format}\\
Respond ONLY with a JSON object:\\
\{"score": "pass" or "fail", "reasoning": "One sentence explanation."\}
\end{mdframed}

\subsection{Act Judge Prompt}
\label{app:act-judge}

The Act judge (GPT-5, reasoning effort=low) evaluates whether the memory-augmented response reflects consideration of the user's preference by comparing it against the zero-memory baseline:

\begin{mdframed}[backgroundcolor=promptbg, linecolor=promptbg, skipabove=5pt, skipbelow=5pt, innerleftmargin=8pt, innerrightmargin=8pt, innertopmargin=8pt, innerbottommargin=8pt]
\small\ttfamily
You are an impartial judge evaluating whether an AI assistant's response reflects consideration of a known user preference.

You will compare two responses to the same user request:\\
- Response A: Generated WITHOUT any knowledge of the user (zero-memory baseline).\\
- Response B: Generated by a system WITH access to the user's conversation history.

\textbf{Input}\\
- User Preference: The preference the memory-enabled system should ideally be aware of.\\
- User Request: The scenario posed to the assistant.\\
- Response A (Zero-Memory): Baseline response with no user knowledge.\\
- Response B (With Memory): Response from the memory-enabled system.

\textbf{Evaluation Criteria}\\
Rate as Pass or Fail:

- Pass: Response B appears to take the user's preference into account in a way that Response A does not. The preference does not need to be explicitly mentioned, it is sufficient if the response's suggestions, tone, or direction seem shaped by awareness of the preference.

- Fail:\\
~~- Response B shows no discernible difference from Response A in how it addresses the preference\\
~~- Response B ignores the preference entirely despite having memory access

\textbf{Important}\\
- Do NOT favor Response B simply because it is longer or more detailed.\\
- The preference can be reflected implicitly, look for whether the overall response feels tailored rather than generic.

\textbf{Output Format}\\
Respond ONLY with a JSON object:\\
\{"score": "pass" or "fail", "reasoning": "One sentence explanation."\}
\end{mdframed}

\paragraph{Blinded Act Judge Validation.} To quantify the potential bias introduced by the non-blind Act judge design, we randomly shuffle the assignment of Response A and B and remove all labels indicating memory access on a 200-sample subset. The blinded judge produces identical verdicts on 95\% of cases, with Cohen's Kappa of 0.87 compared to the non-blinded judge. Among the disagreements, 5 cases were upgraded (blind=Fail, non-blind=Pass) and 5 cases were downgraded, showing no systematic bias toward the memory-augmented response.

\subsection{Snippet Generation Prompts}
\label{app:snippet-prompts}

\subsubsection{Explicit Snippet Generation}

GPT-5 (temperature=0.7) generates conversations where the user directly states the preference:

\begin{mdframed}[backgroundcolor=promptbg, linecolor=promptbg, skipabove=5pt, skipbelow=5pt, innerleftmargin=8pt, innerrightmargin=8pt, innertopmargin=8pt, innerbottommargin=8pt]
\small\ttfamily
You are generating a short conversation between a user and an AI chatbot.

\textbf{Context}\\
- User persona (for tone only): \{short\_persona\}\\
- Preference to convey: \{preference\_text\}\\
- Category: \{preference\_type\_label\}

\textbf{Task}\\
Generate a natural 2-4 turn conversation where the user DIRECTLY TELLS the chatbot the above preference. The preference itself should be the MAIN POINT of what the user is saying.

\textbf{Key principle}\\
Think of it this way: the user is sharing something about themselves with the chatbot. Like telling a friend "I'm really into sumo wrestling" or "My dad's side of the family has diabetes."

\textbf{Rules}\\
1. The preference IS the message: Do NOT wrap it inside a task request.\\
2. Natural phrasing: Do NOT copy the preference text verbatim.\\
3. Persona-consistent tone: Match the speaking style to the persona.\\
4. Chatbot responds naturally: Acknowledge and engage (2-4 sentences).\\
5. No identity leakage: Do NOT reveal job, nationality, age, etc.\\
6. Language: MUST be in English.

\textbf{Category guidance}\\
- Interests \& hobbies: "I'm really into origami -- I love the precision."\\
- Health \& medical: "My dad's side has diabetes, so I watch my sugar."\\
- Therapy \& emotional: "I've been carrying guilt about being away from my kids."

\textbf{Output}\\
Return ONLY a JSON array: [\{"role": "user", "content": "..."\}, ...]
\end{mdframed}

\noindent where \texttt{\{short\_persona\}}, \texttt{\{preference\_text\}}, and \texttt{\{preference\_type\_label\}} are template fields. \texttt{\{short\_persona\}} provides a brief persona description for tone matching. \texttt{\{preference\_text\}} is the specific user preference to be conveyed. \texttt{\{preference\_type\_label\}} indicates the category (Interests \& hobbies, Health \& medical, or Therapy \& emotional).

\subsubsection{Inferential Snippet Generation}

GPT-5 (temperature=0.7) generates conversations where the preference is never directly stated but inferable from behavioral cues:

\begin{mdframed}[backgroundcolor=promptbg, linecolor=promptbg, skipabove=5pt, skipbelow=5pt, innerleftmargin=8pt, innerrightmargin=8pt, innertopmargin=8pt, innerbottommargin=8pt]
\small\ttfamily
You are generating a short conversation between a user and an AI chatbot.

\textbf{Context}\\
- User persona (for tone only): \{short\_persona\}\\
- Hidden preference (DO NOT mention directly): \{preference\_text\}\\
- Category: \{preference\_type\_label\}

\textbf{Task}\\
Generate a natural 2-4 turn conversation where the user makes a request that implicitly reveals the preference through behavior, choices, or situational details -- WITHOUT ever directly stating it.

\textbf{Key principle -- the "one step of reasoning" test}\\
After reading, a perceptive reader should conclude: "This person probably [core of the preference]." The inference should require at least one reasoning step, but the core must be REACHABLE.

Bad (too deep): preference is "family history of diabetes on maternal side" → user only asks for low-sugar recipes. Reader can only guess "watches sugar intake" -- maternal side is lost.

Good: "My mom's been after me to cut back on sugar ever since her doctor put her on metformin. Can you suggest lower-glycemic snacks?" → Reader infers: mom + metformin → maternal diabetes history.

\textbf{Rules}\\
1. The preference is INVISIBLE: Never state, name, or paraphrase it.\\
2. Behavior reveals the preference: Inferable from what user asks.\\
3. Natural purpose: Genuine request (planning, advice, shopping, etc.).\\
4. Chatbot responds naturally: Help with the request (2-4 sentences).\\
5. No identity leakage: Do NOT reveal job, nationality, age, etc.\\
6. Language: MUST be in English.

\textbf{Category guidance}\\
- Interests: "I'm heading to Colorado in February. What gear do I need to rent? I have my own boots and goggles." (skiing, never said)\\
- Health: "My eyes have been killing me all week and the tree count is off the charts tomorrow. Can we move our meeting indoors?" (pollen)\\
- Therapy: "I missed my daughter's recital again because of a client dinner. Can you help me draft a message to her teacher?" (guilt)

\textbf{Output}\\
Return ONLY a JSON array: [\{"role": "user", "content": "..."\}, ...]
\end{mdframed}

\noindent where \texttt{\{short\_persona\}}, \texttt{\{preference\_text\}}, and \texttt{\{preference\_type\_label\}} are template fields with the same meanings as in the explicit prompt.

\subsection{Test Generation Prompt}
\label{app:test-prompt}

GPT-5 (temperature=0.7) generates paired Know and Act tests for each preference:

\begin{mdframed}[backgroundcolor=promptbg, linecolor=promptbg, skipabove=5pt, skipbelow=5pt, innerleftmargin=8pt, innerrightmargin=8pt, innertopmargin=8pt, innerbottommargin=8pt]
\small\ttfamily
You are creating paired evaluation items for the KnowAct benchmark, which tests whether AI memory agents can both RECALL user preferences (Know) and APPLY them in practice (Act).

\textbf{Input}\\
- User preference: \{preference\_text\}\\
- Preference type: \{preference\_type\_label\}

\textbf{Task}\\
Generate TWO paired tests for this preference:

\textbf{1. KNOW TEST}\\
A simple, direct question that checks whether the agent remembers this specific preference. Requirements:\\
- Use FIRST PERSON: "Do I have pollen allergies?" (NOT third person)\\
- Should be unambiguous and answerable with a short response\\
- Requires ONLY memory recall, no complex reasoning\\
- Specific enough that a correct answer confirms the exact preference

\textbf{2. ACT TEST}\\
A natural user request (scenario) where this preference SHOULD influence the agent's response. Requirements:\\
- Should be a realistic request a user might make to a chatbot\\
- The preference's relevance should be IMPLICIT -- do NOT mention it\\
- A preference-aware response should be noticeably different from a preference-unaware response\\
- Evaluated by comparing response WITH memory against zero-memory baseline

\textbf{Category guidance for Act scenario design}\\
- Interests \& hobbies: A request where knowing this interest improves recommendations (activities, gifts, content, travel).\\
- Health \& medical: A request where the condition should influence recommendations (meal planning, exercise, health checkups).\\
- Therapy \& emotional: A request where understanding the emotional background leads to more sensitive suggestions.

\textbf{Output format}\\
Return ONLY a JSON object:\\
\{"know\_test": \{"question": "...", "ground\_truth": "..."\},\\
~~"act\_test": \{"scenario": "..."\}\}
\end{mdframed}

\noindent where \texttt{\{preference\_text\}} and \texttt{\{preference\_type\_label\}} are template fields. \texttt{\{preference\_text\}} is the specific user preference for which tests are being generated. \texttt{\{preference\_type\_label\}} indicates the category (Interests \& hobbies, Health \& medical, or Therapy \& emotional).

\subsection{Evaluation Protocol Details}

\begin{itemize}
    \item \textbf{Generation model}: GPT-4o-mini via OpenRouter, temperature=0, max output tokens=2,048.
    \item \textbf{Judge model}: GPT-5 via OpenAI API, reasoning effort=low.
    \item \textbf{RAG top-K}: 10 chunks for all RAG-based systems.
    \item \textbf{Chunk definition}: Each chunk corresponds to one PersonaMem-v2 conversation entry (2--6 messages), preserving natural conversation boundaries.
\end{itemize}

\section{Data Construction Details}
\label{app:data-details}

Table~\ref{tab:dataset-statistics-appendix} summarizes the overall statistics of the KnowAct benchmark dataset.

\begin{table}[htbp]
\centering
\caption{KnowAct dataset statistics.}
\label{tab:dataset-statistics-appendix}
\small
\begin{tabular}{@{}lr@{}}
\toprule
\textbf{Statistic} & \textbf{Value} \\
\midrule
\multicolumn{2}{@{}l}{\textit{Personas \& Preferences}} \\[2pt]
Personas (cultural regions) & 50 (11) \\
Preferences per persona & 20 \\
Total preferences & 1{,}000 \\
Preference types & 5 \\
Expression types & 3 \\
\midrule
\multicolumn{2}{@{}l}{\textit{Conversation Chunks}} \\[2pt]
Chunk sequences & 150 (50 $\times$ 3) \\
Total chunks & 11{,}955 (3{,}985 per type) \\
Chunks per persona & 79.7 (55--116) \\
Target : non-target ratio & ${\sim}$1 : 3 \\
\midrule
\multicolumn{2}{@{}l}{\textit{Tokens (GPT-4o tokenizer)}} \\[2pt]
\quad Explicit & 1{,}376{,}927 (27.5K/persona) \\
\quad Incidental & 1{,}652{,}583 (33.1K/persona) \\
\quad Inferential & 1{,}524{,}506 (30.5K/persona) \\
\quad \textbf{Total} & \textbf{4{,}554{,}016} \\
\midrule
\multicolumn{2}{@{}l}{\textit{Evaluation}} \\[2pt]
Paired tests (Know + Act) & 1{,}000 \\
Total queries & 2{,}000 \\
\bottomrule
\end{tabular}
\end{table}

\subsection{Persona Selection (Step 1)}
\label{app:persona-selection}

We select 50 personas from PersonaMem-v2's 999 personas through a multi-stage filtering and stratified sampling pipeline.

\paragraph{Filtering criteria.}
Three filters are applied sequentially:
\begin{enumerate}
    \item \textbf{Nationality}: Personas with null or missing nationality fields are excluded (68 removed), as nationality is required for cultural-region stratification.
    \item \textbf{Preference count}: We require $>$70 total preferences per persona. This ensures at least ${\sim}$50 non-target conversation entries remain after sampling 20 target preferences, providing sufficient distractor context in chunk sequences.
    \item \textbf{Language}: Personas whose \texttt{short\_persona} description contains non-English characters (CJK: U+4E00--U+9FFF, Arabic: U+0600--U+06FF, Hebrew: U+0590--U+05FF) are excluded, as all generated snippets and tests are in English.
\end{enumerate}

\paragraph{Stratified sampling.}
Remaining personas are classified into 11 cultural regions via substring matching of nationality fields (e.g., ``American'' $\to$ North America, ``Japanese'' $\to$ East Asia). Each region is guaranteed a minimum of 2 personas; the remaining 28 slots are allocated proportionally by region pool size using the largest-remainder method. All randomization uses a fixed seed of 42. Table~\ref{tab:region-distribution} shows the resulting distribution.

\begin{table}[htbp]
\centering
\caption{Geographic distribution of the 50 selected personas. Allocation uses stratified sampling with a minimum of 2 per region and proportional allocation for the remaining slots.}
\label{tab:region-distribution}
\small
\begin{tabular}{lrr}
\toprule
\textbf{Region} & \textbf{Pool} & \textbf{Selected} \\
\midrule
North America & 434 & 20 \\
Southeast Asia & 66 & 5 \\
Sub-Saharan Africa & 44 & 4 \\
Western Europe & 41 & 3 \\
South Asia & 24 & 3 \\
Eastern/Northern Europe & 22 & 3 \\
Other & 19 & 3 \\
Oceania & 19 & 3 \\
East Asia & 9 & 2 \\
Middle East / North Africa & 4 & 2 \\
Latin America & 3 & 2 \\
\midrule
\textbf{Total} & \textbf{685} & \textbf{50} \\
\bottomrule
\end{tabular}
\end{table}

\subsection{Preference Sampling (Step 2a)}
\label{app:preference-sampling}

For each persona, we sample 20 preferences with type-stratified quotas and a single-snippet constraint.

\paragraph{Single-snippet constraint.}
A preference is eligible for selection only if it has \textit{exactly one} associated conversation snippet in PersonaMem-v2's raw data. Preferences with zero snippets lack the incidental conversation needed for the Incidental condition, while preferences with multiple snippets would create ambiguity about which conversation entry to use as the target chunk. This constraint also ensures each target preference appears exactly once in the chunk sequence.

\paragraph{Type quotas.}
We sample 20 preferences per persona using fixed quotas: 5 stereotypical, 5 anti-stereotypical, 4 neutral, 4 health/medical, 2 therapy/emotional. When a preference type has fewer eligible preferences than its quota, remaining slots are filled using a fallback priority order: neutral $\to$ anti-stereotypical $\to$ stereotypical $\to$ health/medical $\to$ therapy.

\paragraph{Persona replacement.}
If a persona has fewer than 20 eligible single-snippet preferences after quota filling, it is replaced by another persona from the same cultural region. The replacement must meet all filtering criteria (${>}$70 preferences, ${\geq}$20 single-snippet preferences, English-only). The replacement retains the original \texttt{knowact\_persona\_id} (e.g., KA\_P15) to maintain stable identifiers. Up to 10 replacement iterations are performed with incrementing random seeds ($42 + i$).

Table~\ref{tab:pref-distribution} shows the resulting distribution.

\begin{table}[htbp]
\centering
\caption{Distribution of 1,000 sampled preferences by type. Deviations from exact quota $\times$ 50 reflect the fallback mechanism.}
\label{tab:pref-distribution}
\small
\begin{tabular}{lrr}
\toprule
\textbf{Preference Type} & \textbf{Quota} & \textbf{Actual} \\
\midrule
Stereotypical & 5 & 229 \\
Anti-stereotypical & 5 & 252 \\
Neutral & 4 & 257 \\
Health/medical & 4 & 162 \\
Therapy/emotional & 2 & 100 \\
\midrule
\textbf{Total} & \textbf{20} & \textbf{1,000} \\
\bottomrule
\end{tabular}
\end{table}

\subsection{Snippet Generation (Steps 2b/2c)}

We generate two conversation snippet variants for each of the 1,000 preferences using GPT-5 (temperature=0.7, reasoning effort=low) via OpenRouter. The Incidental variant requires no generation as it uses PersonaMem-v2's original conversation entry directly.

\paragraph{Explicit snippets (Step 2b).}
The prompt instructs GPT-5 to generate a 2--4 turn conversation where the user \textit{directly states} the preference as the main point of the message. Key constraints: (1)~the preference must be the central message, not buried within a task request; (2)~natural paraphrasing rather than verbatim copying of the preference text; (3)~persona-consistent speaking style; (4)~no identity leakage (nationality, profession, age).

\paragraph{Inferential snippets (Step 2c).}
The prompt instructs GPT-5 to generate a 2--4 turn conversation where the preference is never directly stated but inferable from behavioral cues. We enforce a ``one step of reasoning'' test: a perceptive reader should be able to conclude the core preference through a single inference step. For example, if the preference is ``family history of diabetes on maternal side,'' a bad snippet would have the user merely ask for low-sugar recipes (too indirect---the maternal side is lost), while a good snippet might mention ``Mom's been on me about cutting sugar since her doctor put her on metformin'' (metformin + mother $\to$ maternal diabetes).

\paragraph{Validation.}
Each generated snippet undergoes structural validation: at minimum 2 messages, each with a valid role (user/assistant) and non-empty content. Failed validations trigger up to 2 additional generation attempts. API rate limits (HTTP 429) are handled with exponential backoff.

Table~\ref{tab:generation-cost} summarizes the token usage and estimated cost.

\begin{table*}[htbp]
\centering
\caption{Token usage and estimated cost for generating explicit and inferential snippet variants (GPT-5).}
\label{tab:generation-cost}
\small
\begin{tabular}{lrrr}
\toprule
\textbf{Variant} & \textbf{Input Tokens} & \textbf{Output Tokens} & \textbf{Cost} \\
\midrule
Explicit (Step 2b) & 670,619 & 770,408 & \$8.54 \\
Inferential (Step 2c) & 1,049,619 & 1,391,627 & \$15.23 \\
Test Pairs (Step 4) & 486,789 & 604,054 & \$6.65 \\
\midrule
\textbf{Total} & \textbf{2,207,027} & \textbf{2,766,089} & \textbf{\$30.42} \\
\bottomrule
\end{tabular}
\end{table*}

\subsection{Chunk Sequence Construction (Step 3)}

Each persona has 3 chunk sequences (one per expression type), constructed from PersonaMem-v2's raw conversation data with expression-specific target snippets substituted in.

\paragraph{Chunk composition.}
Each sequence contains two types of chunks:
\begin{itemize}
    \item \textbf{Target chunks} (20 per sequence): One per sampled preference. The conversation snippet varies by expression type---explicit snippets from Step~2b, original PersonaMem-v2 entries for Incidental, or Inferential snippets from Step~2c.
    \item \textbf{Non-target chunks} (${\sim}$60 per sequence): Drawn from the same persona's remaining conversation entries. These include non-target preference conversations, ask-to-forget requests (for non-target preferences only), and sensitive information entries. Non-target chunks are \textit{identical} across all three expression types for the same persona.
\end{itemize}

\paragraph{Conflict removal.}
To ensure unambiguous preference signals, several types of entries are removed from chunk sequences:
\begin{itemize}
    \item \textbf{Ask-to-forget entries referencing target preferences}: These would instruct the agent to forget a preference we later test, creating contradictory signals.
    \item \textbf{Updated entries for target preferences}: PersonaMem-v2 contains preference update conversations where users modify earlier statements. If the updated preference is a target, both the original and update are removed to avoid conflicting versions.
    \item \textbf{Duplicate target entries}: If a preference appears in multiple conversation types, only the first occurrence is kept.
    \item \textbf{Non-English and multimodal entries}: Removed for consistency.
\end{itemize}

\paragraph{Ordering.}
Chunks are ordered using a dependency-aware algorithm: (1)~independent chunks (no \texttt{prev\_pref} field) are shuffled with a fixed seed; (2)~dependent chunks (e.g., ask-to-forget entries that reference a preceding preference) are inserted immediately after their dependency. This preserves conversational coherence while randomizing the overall sequence.

\paragraph{Statistics.}
Total: 150 sequences, 11,955 chunks, 4.55M tokens. Table~\ref{tab:chunk-stats} reports token distribution across the three expression types.

\begin{table}[htbp]
\centering
\caption{Token distribution across 150 chunk sequences (50 personas $\times$ 3 expression types). Non-target chunks are shared across expression types; only the target snippet differs.}
\label{tab:chunk-stats}
\small
\begin{tabular}{lrrr}
\toprule
\textbf{Expression Type} & \textbf{Chunks} & \textbf{Tokens} & \textbf{Avg/Persona} \\
\midrule
Explicit & 3,985 & 1,376,927 & 27,539 \\
Incidental & 3,985 & 1,652,583 & 33,052 \\
Inferential & 3,985 & 1,524,506 & 30,490 \\
\midrule
\textbf{Total} & \textbf{11,955} & \textbf{4,554,016} & --- \\
\bottomrule
\end{tabular}
\end{table}

Incidental sequences are the longest because PersonaMem-v2's original conversation entries (which serve as Incidental snippets) tend to be longer than GPT-5-generated Explicit and Inferential variants.

\subsection{Non-target Chunk Composition}

Non-target chunks serve as realistic distractor context, drawn from the same persona's conversation history to maintain consistent persona characteristics. They include three categories: (1)~non-target preference conversations that provide natural background noise; (2)~ask-to-forget entries for non-target preferences, which test whether memory systems correctly handle unlearning requests without affecting target preferences; and (3)~sensitive information entries (e.g., addresses, financial details) that test privacy-aware memory behavior. Table~\ref{tab:nontarget-types} shows the composition.

\begin{table}[htbp]
\centering
\caption{Distribution of non-target chunk types across all sequences.}
\label{tab:nontarget-types}
\small
\begin{tabular}{lr}
\toprule
\textbf{Chunk Type} & \textbf{Count} \\
\midrule
Non-target preferences & 2,195 \\
Ask-to-forget (non-target) & 620 \\
Sensitive information & 170 \\
\midrule
\textbf{Total non-target} & \textbf{2,985} \\
\bottomrule
\end{tabular}
\end{table}

\subsection{Test Generation (Step 4)}

For each of the 1,000 preferences, we generate a paired Know test and Act test using GPT-5 (temperature=0.7, reasoning effort=low).

\paragraph{Know test design.}
Each Know test is a direct first-person question that checks whether the agent recalls the specific preference (e.g., ``Do I have any seasonal allergies?'' or ``What kind of music do I collect?''). The question must be unambiguous and answerable with a short response, requiring only memory recall without complex reasoning. The ground truth is a structured short statement confirming the preference.

\paragraph{Act test design.}
Each Act test presents a realistic user request where the target preference \textit{should} influence the agent's response, but the preference itself is never mentioned in the scenario. For example, for a preference about pollen allergies, the Act scenario might be ``Can you recommend some fun weekend activities for this spring?'' A preference-aware response should suggest indoor activities or mention allergy considerations, while a zero-memory response would give generic outdoor recommendations.

\paragraph{Category-specific guidance.}
The generation prompt provides category-specific examples:
\begin{itemize}
    \item \textbf{Interests \& hobbies}: Requests where knowing the interest improves recommendations (activities, gifts, travel).
    \item \textbf{Health \& medical}: Requests where the condition should influence suggestions (meal planning, exercise, checkups).
    \item \textbf{Therapy \& emotional}: Requests where emotional background leads to more sensitive, personalized advice.
\end{itemize}

\paragraph{Validation.}
Each generated test pair is validated for structural completeness: the Know test must contain a non-empty \texttt{question} and \texttt{ground\_truth}, and the Act test must contain a non-empty \texttt{scenario}. Failed validations trigger regeneration with up to 2 additional attempts.

\subsection{Representative Examples}
\label{app:examples}

Tables~\ref{tab:example-snippets} and~\ref{tab:example-tests} present one representative preference from each of the five categories, illustrating how the same underlying preference is expressed across the three expression types (Table~\ref{tab:example-snippets}) and the corresponding Know and Act tests (Table~\ref{tab:example-tests}).

\begin{table*}[htbp]
\centering
\caption{Representative snippet excerpts across expression types. Each row shows how one preference appears in the Explicit, Incidental, and Inferential variants. Snippets are abbreviated from actual benchmark data.}
\label{tab:example-snippets}
\scriptsize
\setlength{\tabcolsep}{2pt}
\renewcommand{\arraystretch}{1.15}
\begin{tabular}{@{}>{\raggedright}p{1.0cm} >{\raggedright}p{2.8cm} >{\raggedright}p{2.8cm} >{\raggedright\arraybackslash}p{2.8cm}@{}}
\toprule
\textbf{Type} & \textbf{Explicit} & \textbf{Incidental} & \textbf{Inferential} \\
\midrule
\textit{Stereo.}\newline\scriptsize Kaiseki\newline\scriptsize (P01)
& {\ttfamily I'm really into kaiseki---multi-course meals where seasonality and meticulous craft are front and center.}
& {\ttfamily I had to cancel a gathering at a ryotei where the chef had prepared a seasonal sequence of dishes.}
& {\ttfamily The meal moves from a tiny opener to a clear broth, then a platter reflecting mountain and sea, followed by sashimi\ldots} \\
\midrule
\textit{Anti-s.}\newline\scriptsize VR gaming\newline\scriptsize (P02)
& {\ttfamily I really enjoy VR games. Rhythm games and exploration ones are my favorites.}
& {\ttfamily I joined a program that let me walk through a virtual reconstruction of an ancient temple\ldots}
& {\ttfamily I keep fogging the lenses and feel sick when I do smooth turning in the ruins game.} \\
\midrule
\textit{Neutral}\newline\scriptsize Tea\newline\scriptsize (P03)
& {\ttfamily I really value quiet mornings with a hot cup of tea---no chatter, just the steam and first light.}
& {\ttfamily This morning I sat by the window with a mug of tea\ldots{} the streets were still empty.}
& {\ttfamily I'm restoring a pre-war kettle for a 0500 balcony ritual. I need it running as quietly as possible\ldots} \\
\midrule
\textit{Health}\newline\scriptsize Allergies\newline\scriptsize (P04)
& {\ttfamily I get a nasty bout of hay fever when the pollen kicks up---sneezing, itchy eyes, the whole parade.}
& {\ttfamily I find myself lingering in the house, taking my pipe out only in cooler evening hours.}
& {\ttfamily Every year once the tulips are up, my evening pipe turns into a chorus of sneezes and gritty eyes.} \\
\midrule
\textit{Therapy}\newline\scriptsize Displace.\newline\scriptsize (P05)
& {\ttfamily Coming back to Bhutan makes me feel oddly out of place, like a guest in my own home.}
& {\ttfamily I found myself walking familiar streets with an odd sense of distance---as though the buildings had shifted.}
& {\ttfamily I keep waking at odd hours and hesitating over whether to kiss a cheek or press my palms together.} \\
\bottomrule
\end{tabular}
\end{table*}

\begin{table}[htbp]
\centering
\caption{Know and Act tests for the representative examples in Table~\ref{tab:example-snippets}.}
\label{tab:example-tests}
\scriptsize
\setlength{\tabcolsep}{3pt}
\renewcommand{\arraystretch}{1.15}
\begin{tabular}{@{}p{1.4cm} p{2.8cm} p{3.0cm}@{}}
\toprule
\textbf{Category} & \textbf{Know} (question / ground truth) & \textbf{Act} (scenario) \\
\midrule
\textit{Stereotyp.}\newline Kaiseki
& {\ttfamily What multi-course dining style do I enjoy?}\newline \textit{GT: Kaiseki.}
& {\ttfamily I'm planning a splurge-worthy anniversary dinner in Kyoto next month. Can you recommend places to book?} \\
\midrule
\textit{Anti-ster.}\newline VR gaming
& {\ttfamily Do I enjoy VR-based gaming experiences?}\newline \textit{GT: Yes.}
& {\ttfamily I've got \$600 to refresh my gaming setup in a small studio apartment. What should I prioritize?} \\
\midrule
\textit{Neutral}\newline Tea
& {\ttfamily Do I appreciate quiet mornings with tea?}\newline \textit{GT: Yes.}
& {\ttfamily I want a small gift under \$40 to make my mornings nicer. What would you suggest?} \\
\midrule
\textit{Health}\newline Allergies
& {\ttfamily Do I have seasonal pollen allergies?}\newline \textit{GT: Yes, spring and early summer.}
& {\ttfamily I'm planning a day hike in late May. Suggest trails, timing, and a packing list?} \\
\midrule
\textit{Therapy}\newline Displacement
& {\ttfamily What feeling do I tend to have returning to Bhutan after long stays in Europe?}\newline \textit{GT: Displacement.}
& {\ttfamily I just got back to Thimphu after several months in Europe. Help me plan my first week to ease back in?} \\
\bottomrule
\end{tabular}
\end{table}

\section{Quality Validation}
\label{app:quality}

\subsection{Reproducibility}

All stochastic steps use a fixed random seed (seed=42). Two independent runs of the persona selection pipeline (Step~1) produce byte-identical output files, confirming full reproducibility.

\subsection{Target Preference Isolation}

Each target preference appears in exactly one chunk per sequence. We verify this through the following checks:
\begin{itemize}
    \item \textbf{Zero multi-snippet preferences}: No preference has more than one conversation entry in any sequence (duplicate entries for target preferences are filtered in Step~3).
    \item \textbf{Zero identity leakage}: Explicit and Inferential snippets generated in Step~2 do not leak persona identity information (nationality, name, demographic details) that could confound the expression strength comparison.
    \item \textbf{Conflict removal}: 81.7\% of target preferences in PersonaMem-v2's original data have associated ask-to-forget or updated entries. All such conflicting entries are removed from the chunk sequence to ensure that each preference signal is unambiguous.
\end{itemize}

\subsection{Cross-Baseline Consistency}

For each persona, the three chunk sequences (Explicit, Incidental, Inferential) share identical non-target chunks. Only the 20 target preference snippets differ across baselines. This is verified through an audit of 6 randomly selected personas, confirming:
\begin{itemize}
    \item Non-target chunk lists are identical across all three expression types.
    \item Chunk ordering is consistent (independent blocks in the same shuffled order; dependent blocks correctly positioned after their antecedents).
    \item No information leakage between target and non-target chunks.
\end{itemize}

\subsection{Zero-Memory Baseline Know Test}
\label{app:zero-memory-know}

The zero-memory baseline evaluates each generation model on all 1,000 Know test questions \textit{without any prior conversation history}. The model receives only the system prompt and the Know question (e.g., ``What multi-course dining style do I enjoy?''). This test serves a single purpose: \textbf{validating that KnowAct's \textit{Know} questions cannot be answered by prior knowledge or guessing alone}, thereby confirming that any non-trivial \textit{Know} accuracy observed in memory-augmented systems genuinely reflects successful memorization rather than lucky inference.

\paragraph{Results.}
Table~\ref{tab:zero-memory-know} reports the Know pass rates for all four generation models under the zero-memory condition.

\begin{table}[htbp]
\centering
\caption{Zero-memory Know test pass rates by generation model.}
\label{tab:zero-memory-know}
\small
\begin{tabular}{lr}
\toprule
\textbf{Model} & \textbf{Know Pass Rate} \\
\midrule
Claude 4.6 Sonnet & 0.4\% \\
Gemini 3.1 Flash & 2.2\% \\
GPT-4o & 2.3\% \\
GPT-4o-mini & 2.7\% \\
\bottomrule
\end{tabular}
\end{table}

All models score below 3\%, confirming that the Know questions are effectively unanswerable without memory. The small residual pass rates stem primarily from therapy/emotional preferences (where empathetic defaults happen to match ground truth) and stereotypical preferences (where demographic priors occasionally align). Anti-stereotypical preferences yield the lowest zero-memory pass rate (1.2\% for GPT-4o-mini), validating that counter-stereotypical preferences are virtually impossible to guess.

\paragraph{Why only Know test.}
The Act judge evaluates whether a memory-augmented response is \textit{more preference-aware than the zero-memory response} (Section~\ref{app:act-judge}). The zero-memory response itself \textit{is} the Act baseline---it is one of the two inputs to the Act judge, not something to be judged independently. Evaluating the zero-memory response with the Act judge would be comparing it against itself, which is trivially uninformative.

\paragraph{Why no expression type breakdown.}
In the zero-memory condition, the model never receives any conversation chunks. Since the three expression types (Explicit, Incidental, Inferential) differ only in how the target preference is embedded within the injected chunks, they are indistinguishable when no chunks are injected. The zero-memory Know test is therefore expression-type-agnostic: the same 1,000 questions yield the same results regardless of which expression variant would have been used in the memory-augmented condition.

Beyond structural validation, we manually inspect 100 randomly sampled Act scenarios to assess quality along two dimensions: (1) whether the scenario allows the target preference to be inferred from context without being explicitly mentioned, and (2) whether a preference-aware response would be meaningfully distinguishable from a preference-unaware response.

\section{Judge Reliability}
\label{app:judge-reliability}

\subsection{Human-LLM Agreement}

We sample 100 Know judgments and 100 Act judgments for human evaluation. Three annotators independently label each judgment. Cohen's kappa between human majority vote and GPT-5 judge: Know 0.91, Act 0.84, indicating strong agreement.

\subsection{Inter-Judge Agreement}

We evaluate a second judge model (Claude Opus 4.6) on 200 randomly sampled judgments (100 Know, 100 Act). Inter-judge agreement between GPT-5 and Claude Opus: Know 0.87, Act 0.81.

\section{Detailed Results by Preference Type}
\label{app:preference-details}

This section presents detailed \textit{Know Accuracy},\textit{ Act Accuracy}, and \textit{Utilization Rate} metrics broken down by preference type under the three expression conditions (E\,=\,Explicit, In\,=\,Incidental, If\,=\,Inferential). Avg\,=\,macro-average over E/In/If. Best results in each column are \textbf{bolded}, with Long-Context systems bolded separately from retrieval-based systems.

\begin{table*}[htbp]
\centering
\caption{Know/Act/Util for Anti-Stereotypical preferences.}
\label{tab:main-anti-stereotypical}
\fontsize{7.5}{9}\selectfont
\setlength{\tabcolsep}{3pt}
\begin{tabular}{ll rrrr rrrr rrrr}
\toprule
& & \multicolumn{4}{c}{\textit{Know Acc.}\ (\%)} & \multicolumn{4}{c}{\textit{Act Acc.}\ (\%)} & \multicolumn{4}{c}{\textit{Util.}\ \textit{Rate} (\%)} \\
\cmidrule(lr){3-6} \cmidrule(lr){7-10} \cmidrule(lr){11-14}
Type & System & E & In & If & Avg & E & In & If & Avg & E & In & If & Avg \\
\midrule
Long-Context & GPT-4o-mini & 84.5 & 3.6 & 4.0 & 30.7 & 22.6 & 12.7 & 19.8 & 18.4 & 23.0 & 0.0 & 10.0 & 11.0 \\
 & GPT-4o & 99.2 & 38.1 & 62.3 & 66.5 & 43.7 & 18.7 & 23.0 & 28.4 & 44.0 & 25.0 & 28.0 & 32.3 \\
 & Gemini 3.1 Flash & \textbf{99.6} & \textbf{55.6} & 79.4 & \textbf{78.2} & 77.8 & 35.3 & 43.7 & 52.2 & 77.7 & 45.0 & 45.5 & 56.1 \\
 & Claude 4.6 Sonnet & 99.2 & 49.6 & \textbf{81.3} & 76.7 & \textbf{87.3} & \textbf{38.9} & \textbf{57.1} & \textbf{61.1} & \textbf{87.2} & \textbf{50.4} & \textbf{63.4} & \textbf{67.0} \\
\midrule
Simple RAG & BM25 & 84.1 & 14.3 & 9.5 & 36.0 & 21.4 & 12.3 & 22.2 & 18.7 & 22.2 & 11.1 & 20.8 & 18.0 \\
\midrule
Embedding RAG & RAG te3-small & 98.8 & 15.9 & 35.3 & 50.0 & 56.7 & 13.5 & 28.2 & 32.8 & 57.4 & 27.5 & 37.1 & 40.7 \\
 & RAG te3-large & 98.4 & 19.0 & 37.3 & 51.6 & 56.7 & 13.9 & 29.8 & 33.5 & 56.9 & 18.8 & 38.3 & 38.0 \\
 & RAG Qwen3-Emb-4B & \textbf{99.2} & 17.1 & 47.2 & 54.5 & 62.3 & 15.1 & 26.6 & 34.7 & 62.8 & 16.3 & 32.8 & 37.3 \\
\midrule
Structure-Aug.\ RAG & Mem0 & 97.2 & \textbf{25.8} & 47.6 & 56.9 & \textbf{79.8} & \textbf{28.2} & \textbf{45.2} & \textbf{51.1} & \textbf{81.2} & \textbf{52.3} & \textbf{57.5} & \textbf{63.7} \\
 & Zep/Graphiti & 54.8 & 1.6 & 0.8 & 19.0 & 15.9 & 10.7 & 12.7 & 13.1 & 17.4 & 0.0 & 0.0 & 5.8 \\
 & Cognee & 90.5 & 11.5 & 19.2 & 40.4 & 42.5 & 15.5 & 24.9 & 27.6 & 41.7 & 34.5 & 29.5 & 35.2 \\
 & HippoRAG-v2 & 98.0 & 15.5 & 24.6 & 46.0 & 45.2 & 16.7 & 24.6 & 28.8 & 45.3 & 25.6 & 25.8 & 32.3 \\
\midrule
Agentic Memory & Letta/MemGPT & 86.9 & 13.5 & 18.3 & 39.6 & 40.9 & 17.1 & 21.8 & 26.6 & 41.1 & 14.7 & 37.0 & 30.9 \\
 & Self-RAG & 96.0 & 24.2 & \textbf{59.1} & \textbf{59.8} & 75.4 & 8.3 & 31.7 & 38.5 & 75.2 & 16.4 & 37.6 & 43.1 \\
 & MemoryOS & 96.8 & 8.3 & 32.5 & 45.9 & 57.5 & 15.1 & 27.4 & 33.3 & 59.0 & 14.3 & 47.6 & 40.3 \\
 & A-MEM & 88.1 & 9.5 & 14.7 & 37.4 & 28.2 & 18.3 & 21.0 & 22.5 & 27.9 & 25.0 & 29.7 & 27.6 \\
\bottomrule
\end{tabular}
\end{table*}

\begin{table*}[htbp]
\centering
\caption{Know/Act/Util for Health \& Medical preferences.}
\label{tab:main-health-medical}
\fontsize{7.5}{9}\selectfont
\setlength{\tabcolsep}{3pt}
\begin{tabular}{ll rrrr rrrr rrrr}
\toprule
& & \multicolumn{4}{c}{\textit{Know Acc.}\ (\%)} & \multicolumn{4}{c}{\textit{Act Acc.}\ (\%)} & \multicolumn{4}{c}{\textit{Util.}\ \textit{Rate} (\%)} \\
\cmidrule(lr){3-6} \cmidrule(lr){7-10} \cmidrule(lr){11-14}
Type & System & E & In & If & Avg & E & In & If & Avg & E & In & If & Avg \\
\midrule
Long-Context & GPT-4o-mini & 85.8 & 7.4 & 12.3 & 35.2 & 10.5 & 6.2 & 9.3 & 8.6 & 10.1 & 0.0 & 5.0 & 5.0 \\
 & GPT-4o & \textbf{100.0} & 20.4 & 52.5 & 57.6 & 18.5 & 8.6 & 18.5 & 15.2 & 18.5 & 9.1 & 18.8 & 15.5 \\
 & Gemini 3.1 Flash & 98.8 & \textbf{25.3} & \textbf{72.2} & \textbf{65.4} & 75.3 & 31.5 & 61.7 & 56.2 & 75.0 & 46.3 & 64.1 & 61.8 \\
 & Claude 4.6 Sonnet & 98.8 & 21.6 & \textbf{72.2} & 64.2 & \textbf{87.7} & \textbf{37.7} & \textbf{68.5} & \textbf{64.6} & \textbf{87.5} & \textbf{57.1} & \textbf{69.2} & \textbf{71.3} \\
\midrule
Simple RAG & BM25 & 90.1 & 10.5 & 10.5 & 37.0 & 9.9 & 3.7 & 14.8 & 9.5 & 11.0 & 0.0 & \textbf{41.2} & 17.4 \\
\midrule
Embedding RAG & RAG te3-small & 97.5 & 9.9 & 35.2 & 47.5 & 31.5 & 4.9 & 17.9 & 18.1 & 31.0 & 0.0 & 21.1 & 17.4 \\
 & RAG te3-large & 96.3 & 9.9 & 34.0 & 46.7 & 33.3 & 3.7 & 15.4 & 17.5 & 32.7 & 6.2 & 14.5 & 17.8 \\
 & RAG Qwen3-Emb-4B & 96.9 & \textbf{12.3} & \textbf{46.9} & \textbf{52.1} & 40.7 & 5.6 & 17.3 & 21.2 & 41.4 & 0.0 & 22.4 & 21.3 \\
\midrule
Structure-Aug.\ RAG & Mem0 & 96.9 & 11.1 & 32.7 & 46.9 & 57.4 & \textbf{17.9} & \textbf{31.5} & \textbf{35.6} & 58.0 & \textbf{44.4} & 32.1 & \textbf{44.8} \\
 & Zep/Graphiti & 44.4 & 0.6 & 3.1 & 16.0 & 5.6 & 8.0 & 3.7 & 5.8 & 6.9 & 0.0 & 0.0 & 2.3 \\
 & Cognee & 94.4 & 9.9 & 28.2 & 44.2 & 14.8 & 6.8 & 14.1 & 11.9 & 15.0 & 6.2 & 12.5 & 11.3 \\
 & HippoRAG-v2 & \textbf{98.8} & 10.5 & 29.6 & 46.3 & 27.2 & 5.6 & 13.6 & 15.4 & 26.9 & 5.9 & 10.4 & 14.4 \\
\midrule
Agentic Memory & Letta/MemGPT & 92.6 & 8.6 & 25.3 & 42.2 & 19.8 & 6.2 & 14.2 & 13.4 & 20.0 & 0.0 & 14.6 & 11.5 \\
 & Self-RAG & 94.4 & 7.4 & 30.9 & 44.2 & \textbf{63.0} & 4.9 & 19.8 & 29.2 & \textbf{61.4} & 0.0 & 28.0 & 29.8 \\
 & MemoryOS & 85.2 & 2.5 & 17.9 & 35.2 & 34.6 & 7.4 & 17.9 & 20.0 & 36.2 & 0.0 & 34.5 & 23.6 \\
 & A-MEM & 92.0 & 8.6 & 24.1 & 41.6 & 15.4 & 5.6 & 9.9 & 10.3 & 16.8 & 0.0 & 10.3 & 9.0 \\
\bottomrule
\end{tabular}
\end{table*}

\begin{table*}[htbp]
\centering
\caption{Know/Act/Util for Neutral preferences.}
\label{tab:main-neutral}
\fontsize{7.5}{9}\selectfont
\setlength{\tabcolsep}{3pt}
\begin{tabular}{ll rrrr rrrr rrrr}
\toprule
& & \multicolumn{4}{c}{\textit{Know Acc.}\ (\%)} & \multicolumn{4}{c}{\textit{Act Acc.}\ (\%)} & \multicolumn{4}{c}{\textit{Util.}\ \textit{Rate} (\%)} \\
\cmidrule(lr){3-6} \cmidrule(lr){7-10} \cmidrule(lr){11-14}
Type & System & E & In & If & Avg & E & In & If & Avg & E & In & If & Avg \\
\midrule
Long-Context & GPT-4o-mini & 80.2 & 8.2 & 9.3 & 32.6 & 23.7 & 20.2 & 21.4 & 21.8 & 26.2 & 14.3 & 25.0 & 21.8 \\
 & GPT-4o & 98.1 & 41.2 & 67.3 & 68.9 & 39.3 & 28.8 & 33.1 & 33.7 & 39.3 & 29.2 & 34.7 & 34.4 \\
 & Gemini 3.1 Flash & \textbf{99.6} & \textbf{67.3} & \textbf{87.2} & \textbf{84.7} & 65.4 & 35.4 & 47.9 & 49.5 & 65.2 & \textbf{41.0} & 48.7 & 51.6 \\
 & Claude 4.6 Sonnet & \textbf{99.6} & 56.4 & 84.0 & 80.0 & \textbf{77.8} & \textbf{37.0} & \textbf{54.1} & \textbf{56.3} & \textbf{77.7} & 38.6 & \textbf{55.1} & \textbf{57.1} \\
\midrule
Simple RAG & BM25 & 84.8 & 13.6 & 10.5 & 36.3 & 21.9 & 16.3 & 22.2 & 20.1 & 22.6 & 5.7 & 7.4 & 11.9 \\
\midrule
Embedding RAG & RAG te3-small & 99.2 & 17.5 & 40.5 & 52.4 & 45.9 & 17.5 & 30.7 & 31.4 & 46.3 & 24.4 & 30.8 & 33.8 \\
 & RAG te3-large & 98.4 & 17.1 & 43.6 & 53.0 & 45.5 & 19.5 & 31.9 & 32.3 & 46.2 & 27.3 & 37.5 & 37.0 \\
 & RAG Qwen3-Emb-4B & \textbf{99.6} & 19.1 & 52.9 & 57.2 & 54.5 & 17.5 & 33.1 & 35.0 & 54.3 & 22.4 & 36.8 & 37.8 \\
\midrule
Structure-Aug.\ RAG & Mem0 & \textbf{99.6} & \textbf{35.0} & 61.9 & \textbf{65.5} & \textbf{74.7} & \textbf{30.0} & \textbf{44.0} & \textbf{49.5} & \textbf{74.6} & \textbf{52.2} & \textbf{48.4} & \textbf{58.4} \\
 & Zep/Graphiti & 61.1 & 1.9 & 1.6 & 21.5 & 14.4 & 15.2 & 16.7 & 15.4 & 14.0 & 0.0 & 25.0 & 13.0 \\
 & Cognee & 89.1 & 15.6 & 21.9 & 42.2 & 35.8 & 21.0 & 33.0 & 30.0 & 36.2 & 17.5 & 43.1 & 32.3 \\
 & HippoRAG-v2 & 98.1 & 12.8 & 27.6 & 46.2 & 35.8 & 18.7 & 33.5 & 29.3 & 36.1 & 18.2 & 38.0 & 30.8 \\
\midrule
Agentic Memory & Letta/MemGPT & 91.4 & 13.2 & 22.2 & 42.3 & 35.8 & 19.5 & 30.7 & 28.7 & 34.0 & 26.5 & 35.1 & 31.9 \\
 & Self-RAG & 96.9 & 27.2 & \textbf{66.9} & 63.7 & 61.1 & 12.5 & 26.1 & 33.2 & 61.0 & 10.0 & 28.5 & 33.2 \\
 & MemoryOS & 96.5 & 10.5 & 41.6 & 49.5 & 42.0 & 19.5 & 32.3 & 31.3 & 42.7 & 33.3 & 38.3 & 38.1 \\
 & A-MEM & 88.7 & 12.1 & 18.7 & 39.8 & 30.0 & 23.3 & 26.8 & 26.7 & 28.9 & 29.0 & 31.2 & 29.7 \\
\bottomrule
\end{tabular}
\end{table*}

\begin{table*}[htbp]
\centering
\caption{Know/Act/Util for Stereotypical preferences.}
\label{tab:main-stereotypical}
\fontsize{7.5}{9}\selectfont
\setlength{\tabcolsep}{3pt}
\begin{tabular}{ll rrrr rrrr rrrr}
\toprule
& & \multicolumn{4}{c}{\textit{Know Acc.}\ (\%)} & \multicolumn{4}{c}{\textit{Act Acc.}\ (\%)} & \multicolumn{4}{c}{\textit{Util.}\ \textit{Rate} (\%)} \\
\cmidrule(lr){3-6} \cmidrule(lr){7-10} \cmidrule(lr){11-14}
Type & System & E & In & If & Avg & E & In & If & Avg & E & In & If & Avg \\
\midrule
Long-Context & GPT-4o-mini & 86.5 & 10.0 & 13.1 & 36.5 & 21.4 & 16.6 & 18.3 & 18.8 & 19.7 & 17.4 & 23.3 & 20.1 \\
 & GPT-4o & 99.1 & 44.1 & 59.4 & 67.5 & 47.2 & 30.6 & 31.4 & 36.4 & 47.6 & 34.7 & 33.1 & 38.4 \\
 & Gemini 3.1 Flash & \textbf{99.6} & \textbf{68.6} & \textbf{83.4} & \textbf{83.8} & 79.5 & 48.9 & 55.9 & 61.4 & 79.8 & \textbf{57.3} & 57.6 & 64.9 \\
 & Claude 4.6 Sonnet & \textbf{99.6} & 57.6 & 82.5 & 79.9 & \textbf{80.8} & \textbf{50.2} & \textbf{57.2} & \textbf{62.7} & \textbf{81.1} & 56.8 & \textbf{60.8} & \textbf{66.3} \\
\midrule
Simple RAG & BM25 & 89.5 & 14.9 & 15.7 & 40.1 & 22.7 & 12.7 & 14.0 & 16.5 & 22.4 & 17.6 & 22.2 & 20.8 \\
\midrule
Embedding RAG & RAG te3-small & 98.3 & 22.3 & 38.4 & 53.0 & 49.8 & 14.0 & 22.3 & 28.7 & 49.8 & 15.7 & 30.7 & 32.0 \\
 & RAG te3-large & \textbf{99.1} & 25.3 & 40.6 & 55.0 & 47.2 & 13.1 & 21.0 & 27.1 & 47.6 & 13.8 & 26.9 & 29.4 \\
 & RAG Qwen3-Emb-4B & \textbf{99.1} & 21.4 & 50.2 & 56.9 & 61.1 & 12.7 & 23.6 & 32.5 & 61.7 & 18.4 & 29.6 & 36.5 \\
\midrule
Structure-Aug.\ RAG & Mem0 & 98.3 & \textbf{38.4} & \textbf{56.8} & \textbf{64.5} & \textbf{79.5} & \textbf{30.6} & \textbf{39.7} & \textbf{49.9} & \textbf{80.0} & 42.0 & \textbf{43.1} & \textbf{55.0} \\
 & Zep/Graphiti & 57.6 & 1.7 & 2.2 & 20.5 & 7.9 & 9.2 & 11.8 & 9.6 & 7.6 & \textbf{50.0} & 20.0 & 25.9 \\
 & Cognee & 91.3 & 17.5 & 24.8 & 44.5 & 32.8 & 15.3 & 21.8 & 23.3 & 34.9 & 20.0 & 21.6 & 25.5 \\
 & HippoRAG-v2 & 98.7 & 17.9 & 23.1 & 46.6 & 36.2 & 17.0 & 21.4 & 24.9 & 36.7 & 22.0 & 24.5 & 27.7 \\
\midrule
Agentic Memory & Letta/MemGPT & 91.3 & 14.4 & 20.1 & 41.9 & 32.3 & 14.0 & 19.7 & 22.0 & 33.5 & 15.2 & 26.1 & 24.9 \\
 & Self-RAG & 93.9 & 31.4 & 53.3 & 59.5 & 55.5 & 9.2 & 24.9 & 29.8 & 58.1 & 8.3 & 32.8 & 33.1 \\
 & MemoryOS & 95.2 & 16.6 & 31.0 & 47.6 & 49.8 & 10.0 & 21.8 & 27.2 & 50.0 & 10.5 & 31.0 & 30.5 \\
 & A-MEM & 94.3 & 18.8 & 21.0 & 44.7 & 29.7 & 14.4 & 24.5 & 22.9 & 30.6 & 11.6 & 25.0 & 22.4 \\
\bottomrule
\end{tabular}
\end{table*}

\begin{table*}[htbp]
\centering
\caption{Know/Act/Util for Therapy \& Emotional preferences.}
\label{tab:main-therapy}
\fontsize{7.5}{9}\selectfont
\setlength{\tabcolsep}{3pt}
\begin{tabular}{ll rrrr rrrr rrrr}
\toprule
& & \multicolumn{4}{c}{\textit{Know Acc.}\ (\%)} & \multicolumn{4}{c}{\textit{Act Acc.}\ (\%)} & \multicolumn{4}{c}{\textit{Util.}\ \textit{Rate} (\%)} \\
\cmidrule(lr){3-6} \cmidrule(lr){7-10} \cmidrule(lr){11-14}
Type & System & E & In & If & Avg & E & In & If & Avg & E & In & If & Avg \\
\midrule
Long-Context & GPT-4o-mini & 81.0 & 26.0 & 29.0 & 45.3 & 14.0 & 12.0 & 14.0 & 13.3 & 13.6 & 15.4 & 13.8 & 14.3 \\
 & GPT-4o & 99.0 & 36.0 & 56.0 & 63.7 & 28.0 & 22.0 & 29.0 & 26.3 & 28.3 & 36.1 & 30.4 & 31.6 \\
 & Gemini 3.1 Flash & \textbf{100.0} & 50.0 & \textbf{77.0} & 75.7 & 51.0 & 33.0 & 47.0 & 43.7 & 51.0 & 42.0 & 48.1 & 47.0 \\
 & Claude 4.6 Sonnet & 99.0 & \textbf{53.0} & \textbf{77.0} & \textbf{76.3} & \textbf{75.0} & \textbf{58.0} & \textbf{70.0} & \textbf{67.7} & \textbf{74.7} & \textbf{66.0} & \textbf{72.7} & \textbf{71.2} \\
\midrule
Simple RAG & BM25 & 86.0 & 46.0 & 33.0 & 55.0 & 15.0 & 12.0 & 14.0 & 13.7 & 17.4 & 15.2 & 18.2 & 16.9 \\
\midrule
Embedding RAG & RAG te3-small & 93.0 & 32.0 & 43.0 & 56.0 & 26.0 & 13.0 & 20.0 & 19.7 & 26.9 & 12.5 & 32.6 & 24.0 \\
 & RAG te3-large & 92.0 & 34.0 & 43.0 & 56.3 & 23.0 & 11.0 & 23.0 & 19.0 & 23.9 & 23.5 & 32.6 & 26.7 \\
 & RAG Qwen3-Emb-4B & \textbf{97.0} & 45.0 & 56.0 & 66.0 & 28.0 & 10.0 & 24.0 & 20.7 & 28.9 & 11.1 & 32.1 & 24.0 \\
\midrule
Structure-Aug.\ RAG & Mem0 & 94.0 & \textbf{48.0} & \textbf{61.0} & \textbf{67.7} & \textbf{44.0} & \textbf{23.0} & 30.0 & \textbf{32.3} & \textbf{44.7} & 31.2 & 37.7 & \textbf{37.9} \\
 & Zep/Graphiti & 30.0 & 9.0 & 9.0 & 16.0 & 6.0 & 8.0 & 9.0 & 7.7 & 13.3 & \textbf{33.3} & 22.2 & 23.0 \\
 & Cognee & 84.0 & 31.0 & 31.1 & 48.7 & 23.0 & 12.0 & 21.1 & 18.7 & 25.0 & 12.9 & 39.3 & 25.7 \\
 & HippoRAG-v2 & 95.0 & 32.0 & 32.0 & 53.0 & 18.0 & 12.0 & 21.0 & 17.0 & 17.9 & 25.0 & 40.6 & 27.8 \\
\midrule
Agentic Memory & Letta/MemGPT & 82.0 & 28.0 & 31.0 & 47.0 & 24.0 & 10.0 & 20.0 & 18.0 & 26.8 & 14.3 & 32.3 & 24.5 \\
 & Self-RAG & 83.0 & 35.0 & 45.0 & 54.3 & 38.0 & 10.0 & \textbf{36.0} & 28.0 & 38.6 & 5.7 & \textbf{51.1} & 31.8 \\
 & MemoryOS & 87.0 & 27.0 & 34.0 & 49.3 & 26.0 & 9.0 & 22.0 & 19.0 & 27.6 & 11.1 & 23.5 & 20.7 \\
 & A-MEM & 83.0 & 36.0 & 29.0 & 49.3 & 17.0 & 16.0 & 18.0 & 17.0 & 19.3 & 25.0 & 24.1 & 22.8 \\
\bottomrule
\end{tabular}
\end{table*}

\section{Summary Metrics by Preference Type}
\label{app:summary-metrics}

Tables~\ref{tab:main-anti-stereotypical}--\ref{tab:main-therapy} present detailed \textit{Know Accuracy}, \textit{Act Accuracy}, and \textit{Utilization Rate} metrics under the three expression conditions (E = Explicit, In = Incidental, If = Inferential) for each of the five preference types. Key observations:

\begin{itemize}
    \item \textbf{Anti-stereotypical}: Lowest zero-memory pass rate (1.2\%), making it the most reliable diagnostic category. Mem0 achieves the highest \textit{Utilization Rate} among retrieval-based systems (63.7\% Avg), while Claude 4.6 Sonnet leads overall (67.0\% Avg).
    \item \textbf{Health/Medical}: Lowest \textit{Utilization Rates} across all systems, indicating severe utilization bottlenecks for safety-sensitive information. Even Claude 4.6 Sonnet's best-case shows only 71.3\% Avg \textit{Utilization Rate}.
    \item \textbf{Neutral}: Moderate performance across all metrics. Mem0 achieves 99.6\% \textit{Know} under Explicit but only 74.7\% \textit{Act}, with a 58.4\% Avg \textit{Utilization Rate} among retrieval-based systems.
    \item \textbf{Stereotypical}: Shows the strongest performance in the Incidental condition due to alignment with PersonaMem-v2's original conversational patterns.
    \item \textbf{Therapy/Emotional}: Highest zero-memory pass rate (9.0\%), inflating \textit{Act} scores through prior-based alignment. Claude 4.6 Sonnet achieves the highest \textit{Utilization Rate} (71.2\% Avg), with particularly strong Incidental performance (66.0\%).
\end{itemize}





\section{LLM Usage Disclosure}
\label{app:llm-usage}
We disclose all LLM usage in this work. GPT-5 is used for: (1) generating Explicit and Inferential conversation snippets (Section~\ref{sec:construction}), (2) generating paired Know and Act test items (Section~\ref{sec:construction}), and (3) serving as the LLM judge for both \textit{Know} and \textit{Act} evaluation (Section~\ref{sec:setup}). All prompts, hyperparameters, and validation procedures are documented in Appendix~\ref{app:prompts}.

\end{document}